\documentclass[sts, twocolumn]{imsart}

\usepackage{amsthm}
\usepackage{amsmath}
\usepackage{natbib}
\usepackage[colorlinks,citecolor=blue,urlcolor=blue,filecolor=blue,backref=page]{hyperref}
\usepackage{graphicx}

\usepackage{algorithm}
\usepackage{algorithmicx}
\usepackage{algpseudocode}
\usepackage{amssymb}
\usepackage{bbm}
\usepackage{booktabs}
\usepackage{hhline}
\usepackage{multirow}
\usepackage{subcaption}
\usepackage[table]{xcolor}

\startlocaldefs
\numberwithin{equation}{section}
\theoremstyle{plain}

\endlocaldefs

\DeclareMathOperator*{\argmax}{\arg\max}

\begin{document}

\begin{frontmatter}

% "Title of the paper"
% \title{Challenges in Bayesian inference via
% Markov chain Monte Carlo for neural networks}
\title{Challenges in Markov chain Monte Carlo for Bayesian neural networks}
\runtitle{Challenges in MCMC for Bayesian neural networks}

% indicate corresponding author with \corref{}
% \author{\fnms{John} \snm{Smith}\corref{}\ead[label=e1]{smith@foo.com}\thanksref{t1}}
% \thankstext{t1}{Thanks to somebody}
% \address{line 1\\ line 2\\ printead{e1}}
% \affiliation{Some University}

% \author{\fnms{???} \snm{???}\ead[label=e1]{???}}
% \address{\printead{e1}}
%\and
%\author{\fnms{???} \snm{???}\ead[label=e2]{???}}
%\address{\printead{e2}}
%\affiliation{???}

\author{\fnms{Theodore} \snm{Papamarkou}\corref{}\ead[label=e1]{
theodoros.papamarkou@manchester.ac.uk}},
\author{\fnms{Jacob} \snm{Hinkle}\ead[label=e2]{
hinklejd@ornl.gov}},
\author{\fnms{M. Todd} \snm{Young}\ead[label=e3]{
youngmt1@ornl.gov}}
\and
\author{\fnms{David} \snm{Womble}\ead[label=e4]{
womblede@ornl.gov}}
\address{Department of Mathematics,
The University of Manchester, Manchester, UK, and
Computational Sciences and Engineering Division, Oak Ridge National
Laboratory, Oak Ridge, TN, USA}
% \address{Theodore Papamarkou is a Reader in the Mathematics of Data Science,
% Department of Mathematics, The University of Manchester, Manchester, UK
% \printead{e1}.
%Jacob Hinkle is a Research Scientist,
%Computational Sciences and Engineering Division, Oak Ridge National
%Laboratory, Oak Ridge, TN, USA \printead{e2}.
%M. Todd Young is a Machine Learning Engineer,
%Computational Sciences and Engineering Division, Oak Ridge National
%Laboratory, Oak Ridge, TN, USA \printead{e3}.
%David Womble is an AI Program Director,
%Computational Sciences and Engineering Division, Oak Ridge National
%Laboratory, Oak Ridge, TN, USA \printead{e4}.}

\runauthor{T. Papamarkou, J. Hinkle, M. T. Young and D. Womble}

\begin{abstract}
Markov chain Monte Carlo (MCMC) methods
have not been broadly adopted in Bayesian neural networks (BNNs).
This paper initially reviews the main challenges in sampling
from the parameter posterior of a neural network via MCMC.
Such challenges culminate to lack of convergence
to the parameter posterior.
Nevertheless, this paper shows that
a non-converged Markov chain, generated via MCMC sampling from
the parameter space of a neural network,
can yield via Bayesian marginalization
a valuable posterior predictive distribution of the output of the neural network.
Classification examples based on multilayer perceptrons showcase
highly accurate posterior predictive distributions.
The postulate of limited scope
for MCMC developments in BNNs is partially valid;
an asymptotically exact parameter posterior seems less plausible,
yet an accurate posterior predictive distribution is a tenable research avenue.
\end{abstract}

%\begin{keyword}[class=MSC]
%\kwd[Primary ]{}
%\kwd{}
%\kwd[; secondary ]{}
%\end{keyword}

%\begin{keyword}
%\kwd{}
%\kwd{}
%\end{keyword}


\end{frontmatter}

\section{Motivation}

The \textit{universal approximation theorem} \citep{cybenko1989}
and its subsequent extensions \citep{hornik1991, lu2017} state that
feedforward neural networks with exponentially large width and
width-bounded deep neural networks can approximate any continuous function
arbitrarily well.
This universal approximation capacity of neural networks
along with available computing power
explain the widespread use of deep learning nowadays.

Bayesian inference for neural networks is typically performed
via stochastic Bayesian optimization,
stochastic variational inference \citep{polson2017}
or ensemble methods \citep{ashukha2020, wilson2020}.
MCMC methods
have been explored in the context of neural networks, but
have not become part of the Bayesian deep learning toolbox.

The slower evolution of MCMC methods for neural networks
is partly attributed to the lack of scalability of existing MCMC algorithms
for big data and for high-dimensional parameter spaces.
Furthermore, additional factors hinder
the adaptation of existing MCMC methods in deep learning,
including the
% non-linear
hierarchical structure of neural networks
and the associated covariance between parameters,
lack of identifiability arising from weight symmetries,
lack of a priori knowledge about the parameter space,
% and the associated impact of non-objective priors on the parameter posterior,
and ultimately lack of convergence.

The purpose of this paper is twofold.
Initially,
a literature review is conducted
to identify inferential challenges
in MCMC developments for neural networks.
% over the last three decades.
Subsequently,
Bayesian marginalization
based on MCMC samples of neural network parameters
is used
for attaining accurate posterior predictive distributions
of the respective neural network output,
despite the lack of convergence
of the MCMC samples to the parameter posterior.

An outline of the paper layout follows.
Section \ref{challenges}
reviews the inferential challenges
arising from the application of MCMC to neural networks.
Section \ref{inferential_overview}
provides an overview
of the employed inferential framework,
including the multilayer perceptron (MLP) model
and its likelihood
for binary and multiclass classification,
the MCMC algorithms for sampling from MLP parameters,
the multivariate MCMC diagnostics
for assessing convergence and sampling effectiveness,
and the Bayesian marginalization
for attaining posterior predictive distributions of MLP outputs.
Section \ref{examples}
showcases Bayesian parameter estimation via MCMC
and Bayesian predictions via marginalization
by fitting different MLPs to four datasets.
Section \ref{scope} posits predictive inference
for neural networks, among else by combining
Bayesian marginalization
with approximate MCMC sampling
or with ensemble training.
% and section \ref{future_work}
% suggests future steps
% towards this research direction.

% \section{Inferential challenges}
\section{Parameter inference challenges}
\label{challenges}

A literature review of inferential challenges
in the application of MCMC methods to neural networks
is conducted in this section thematically,
with each subsection being focused on a different challenge.

\subsection{Computational cost}
\label{computational_cost}

Existing MCMC algorithms do not scale
with increasing number of parameters or of data points.
For this reason, approximate inference methods, including variational inference (VI),
are preferred in high-dimensional parameter spaces or in big data problems
from a time complexity standpoint \citep{mackay1995, blei2017, blier2018}.
On the other hand, MCMC methods are better than VI
in terms of approximating the log-likelihood \citep{dupuy2017}.

Literature on MCMC methods for neural networks
is limited due to associated computational complexity implications.
Sequential Monte Carlo and reversible jump MCMC have been applied
on two types of neural network architectures, namely MLPs and radial basis function networks (RBFs),
see for instance \citet{andrieu1999, freitas1999, andrieu2000, freitas2001}.
For a review of Bayesian approaches to neural networks, see \citet{titterington2004}.

Many research developments have been made to
scale MCMC algorithms to big data.
The main focus has been on designing Metropolis-Hastings or Gibbs sampling variants that evaluate
a costly log-likelihood on a subset (\textit{minibatch}) of the data rather than on the entire data set
\citep{welling2011, chen2014, ma2017, mandt2017,
desa2018, nemeth2018, robert2018, seita2018, quiroz2019}.

Among minibatch MCMC algorithms to big data applications,
there exists a subset of studies
applying such algorithms to neural networks
\citep{chen2014, gu2015, gong2019}.
Minibatch MCMC approaches to neural networks pave the way towards \textit{data-parallel deep learning}.
On the other hand, to the best of the authors' knowledge,
there is no published research on MCMC methods
that evaluate the log-likelihood
on a subset of neural network parameters rather than on the whole set of parameters,
and therefore no reported research on \textit{model-parallel deep learning} via MCMC.

Minibatch MCMC has been studied analytically by \citet{johndrow2020}.
Their theoretical findings point out that
some minibatching schemes can yield inexact approximations
and that minibatch MCMC can not greatly expedite the rate of convergence.

\subsection{Model structure}

A neural network with $\rho$ layers can be viewed as a hierarchical model with $\rho$ levels,
each network layer representing a level \citep{williams2000}.
Due to
its nested layers and
its non-linear activations,
a neural network is
a \textit{non-linear hierarchical model}.

MCMC methods for non-linear hierarchical models have been developed,
see for example \citet{bennett1996, gilks1996b, daniels1998, sargent2000}.
However, existing MCMC methods
for non-linear hierarchical models
have not harnessed neural networks
due to time complexity and convergence implications.

Although not designed to mirror the hierarchical structure of a neural network,
recent hierarchical VI
\citep{ranganath2016, esmaeili2019, huang2019, titsias2019}
provides more general variational
approximations of the parameter posterior of the neural network
than mean-field VI.
Introducing a hierarchical structure in the variational distribution
induces correlation among parameters,
in contrast to the mean-field variational distribution that assumes independent parameters.
% In contrast, the mean-field variational distribution assumes independence
% among networks and biases,
% which yields to a computationally cheaper and
% less faithful representation of the parameter posterior.
So, one of the Bayesian inference strategies for neural networks is to
approximate the covariance structure among network parameters.
In fact, there are published comparisons between MCMC and VI
in terms of speed and accuracy of convergence to the posterior covariance,
both for linear or mixture models \citep{giordano2015, mandt2017, ong2018}
and for neural networks \citep{zhang2018a}.

\subsection{Weight symmetries}

The output of a feedforward neural network
given some fixed input
remains unchanged under a set of transformations determined by the
the choice of activations and
by the network architecture more generally.
For instance,
certain weight permutations and sign flips in
MLPs with hyperbolic tangent activations
leave the output unchanged \citep{chen1993}.

If a parameter transformation leaves the output of a neural network unchanged
given some fixed input,
then the likelihood is invariant under the transformation.
In other words, transformations, such as weight permutations and sign-flips,
render neural networks \textit{non-identifiable} \citep{pourzanjani2017}.

It is known that the set of linear invertible parameter transformations that leaves
the output unchanged is a subgroup $T$ of the group of invertible linear mappings
from the parameter space $\mathbb{R}^n$ to itself \citep{nielsen1990}.
$T$ is a transformation group acting on the parameter space $\mathbb{R}^n$.
It can be shown that for each permutable feedforward neural network,
there exists a cone $H\subset\mathbb{R}^{n}$ dependent only on the network architecture
such that for any parameter $\theta\in\mathbb{R}^n$ there exist
$\eta\in H$ and $\tau\in T$ such that $\tau\eta=\theta$.
This relation means that every network parameter is equivalent
to a parameter in the proper subset $H$ of $\mathbb{R}^n$ \citep{nielsen1990}.
Neural networks with convolutions, max-pooling and batch-normalization
contain more types of weight symmetries than
MLPs \citep{badrinarayanan2015}.

In practice, the parameter space of a neural network is set to be
the whole of $\mathbb{R}^n$ rather than a cone $H$ of $\mathbb{R}^n$.
Since a neural network likelihood with support in
the non-reduced parameter space of $\mathbb{R}^n$
is invariant under weight permutations, sign-flips or other transformations,
the posterior landscape includes multiple equally likely modes.
This implies
low acceptance rate,
entrapment in local modes and
convergence challenges for MCMC.
Additionally, computational time is wasted during MCMC,
since posterior modes represent equivalent solutions \citep{nalisnick2018}.
Such challenges manifest themselves in the MLP examples of section \ref{examples}.
For neural networks with higher number $n$ of parameters in $\mathbb{R}^n$,
the topology of the likelihood
is characterized by local optima embedded
in high-dimensional flat plateaus \citep{brea2019}.
Thereby, larger neural networks lead to
a multimodal target density with symmetric modes for MCMC.

Seeking parameter symmetries in neural networks
can lead to a variety of NP-hard problems \citep{ensign2017}.
Moreover, symmetries in neural networks pose
identifiability and associated inferential challenges in Bayesian inference,
but they also provide opportunities to develop inferential methods
with reduced computational cost \citep{hu2019}
or with improved predictive performance \citep{moore2016}.
Empirical evidence from stochastic optimization simulations
suggests that removing weight symmetries
has a negative effect on prediction accuracy
in smaller and shallower convolutional neural networks (CNNs),
but has no effect in prediction accuracy in larger and deeper CNNs
\citep{maddison2015}.

Imposing constraints on neural network weights
is one way of removing symmetries,
leading to better mixing for MCMC
\citep{sen2020}.
More generally, exploitation of weight symmetries
provides scope for scalable Bayesian inference in deep learning
by reducing the measure or dimension of parameter space.
% with reduced computational cost and without losing prediction accuracy
Bayesian inference in subspaces of parameter space for deep learning
has been proposed before
\citep{izmailov2020}.

Lack of identifiability is not unique to neural networks.
For instance, the likelihood of mixture models
is invariant under relabelling of the mixture components,
a condition known as the \textit{label switching} problem
\citep{stephens2000}.

The high-dimensional parameter space of neural networks
is another source of non-identifiability.
A necessary condition for identifiability is that
the number of data points must be larger
than the number of parameters.
This is one reason  why big datasets are required
for training neural networks.

\subsection{Prior specification}

Parameter priors have been used for generating
Bayesian smoothing or regularization effects.
For instance, \citet{freitas1999} develops
sequential Monte Carlo methods with smoothing priors for MLPs
and \citet{williams1995}
introduces Bayesian regularization and pruning
for neural networks via a Laplace prior.

When parameter prior specification for a neural network
is not driven by smoothing or regularization,
the question becomes how to choose the prior.
The choice of parameter prior for a neural network
is crucial in that it affects the parameter posterior \citep{lee2004},
and consequently the posterior predictive distribution \citep{lee2005}.

Neural networks are commonly applied to big data.
For large amounts of data,
practitioners may not have
intuition about the relationship between input and output variables.
Furthermore, it is an open research question
how to interpret neural network weights and biases.
As a priori knowledge about big datasets and about neural network parameters
is typically not available,
\textit{prior elicitation} from experts is
not applicable to neural networks.
% in practice.

It seems logical to choose a prior that reflects a priori ignorance
about the parameters.
A constant-valued prior is a possible candidate,
with the caveat of being improper for unbounded parameter spaces,
such as $\mathbb{R}^n$.
However, for neural networks,
an \textit{improper prior}
can result in an improper parameter posterior \citep{lee2005}.

Typically, a \textit{truncated flat prior} for neural networks
is sufficient for ensuring a valid parameter posterior \citep{lee2005}.
At the same time, the choice of truncation bounds
depends on weight symmetry and consequently on the allocation of
equivalent points in the parameter space.
\citet{lee2003} proposes a \textit{restricted flat prior}
for feedforward neural networks
by bounding some of the parameters and by imposing constraints that
guarantee layer-wise linear independence between activations,
while \citet{lee2000} shows that this prior is asymptotically consistent
for the posterior.
Moreover, \citet{lee2003} demonstrates that such a restricted flat prior
enables more effective MCMC sampling in comparison to alternative prior choices.

\textit{Objective prior specification} is an area of statistics
that has not infiltrated Bayesian inference for neural networks.
Alternative ideas for constructing objective priors
with minimal effect on posterior inference
exist in the statistics literature.
For example,
\textit{Jeffreys priors}
are invariant to differentiable one-to-one transformations of the parameters \citep{jeffreys1962},
\textit{maximum entropy priors} maximize the Shannon entropy
and therefore provide the least possible information \citep{jaynes1968},
\textit{reference priors}
maximize the expected Kullback-Leibler divergence from the associated posteriors
and in that sense are the least informative priors \citep{bernardo1979},
and \textit{penalised complexity priors}
penalise the complexity induced by deviating from a simpler base model \citep{simpson2017}.

To the best of the authors' knowledge,
there are only two published lines of research on objective priors for neural networks;
a theoretical derivation of Jeffreys and reference priors for feedforward neural networks
by \citet{lee2007},
and an approximation of reference priors
via Monte Carlo sampling of a differentiable non-centered parameterization
of MLPs and CNNs by \citet{nalisnick2018}.

More broadly, research on prior specification for BNNs
has been published recently \citep{pearce2019, vladimirova2019}.
For a more thorough review of prior specification for BNNs, see \citet{lee2005}.

\subsection{Convergence}

MCMC convergence depends on the target density,
namely on its multi-modality and level of smoothness.
An MLP with fewer than a hundred parameters
fitted to a non-linearly separable dataset
makes convergence in fixed MCMC sampling time
challenging (see subsection \ref{num_summaries}).

Attaining MCMC convergence is not the only challenge.
Assessing whether a finite sample from an MCMC algorithm represents
an underlying target density can not be done with certainty \citep{cowles1996}.
MCMC diagnostics can fail to detect the type of convergence failure
they were designed to identify.
Combinations of diagnostics are thus used in practice
to evaluate MCMC convergence
with reduced risk of false diagnosis.
%In this paper,
%the potential scale reduction factor (PSRF) and the effective sample size (ESS)
%are used jointly to assess MCMC convergence
%(see subsection \ref{mcmc_diagnostics}).

MCMC diagnostics were initially designed for
asymptotically exact MCMC.
Research activity on approximate MCMC has emerged recently.
Minibatch MCMC methods (see subsection \ref{computational_cost})
are one class of approximate MCMC methods.
Alternative approximate MCMC techniques without minibatching have been developed
\citep{rudolf2018, chen2019}
along with new approaches to quantify convergence
\citep{chwialkowski2016}.

\textit{Quantization} and \textit{discrepancy} are two notions
pertinent to approximate MCMC methods.
The quantization of a target density $p$ by an empirical measure $\hat{p}$
provides an approximation to the target $p$ \citep{graf2007},
while the notion of discrepancy quantifies
how well the empirical measure $\hat{p}$ approximates the target $p$
\citep{chen2019}.
The \textit{kernel Stein discrepancy} (KSD)
and the \textit{maximum mean discrepancy} (MMD)
constitute two instances of discrepancy;
for more details, see \citet{chen2019} and \citet{gretton2012}, respectively.
\citet{rudolf2018} provide an alternative way of assessing
the quality of approximation of a target density $p$ by
an empirical measure $\hat{p}$
in the context of approximate MCMC
using the notion of \textit{Wasserstein distance} between
$p$ and $\hat{p}$.

\section{Inferential framework overview}
\label{inferential_overview}

An overview of the inferential framework used in this paper follows,
including
the MLP model and its likelihood for classification,
MCMC samplers for parameter estimation,
MCMC diagnostics for assessing convergence and sampling effectiveness,
and Bayesian marginalization for prediction.

\subsection{The MLP model}
\label{mlp_overview}

% Keeping the scope of this study in mind,
MLPs have been chosen as a more tractable class of neural networks.
CNNs are the most widely used deep learning models.
However, even small CNNs,
such as AlexNet \citep{krizhevsky2012},
SqueezeNet \citep{iandola2016},
Xception \citep{chollet2017},
MobileNet \citep{howard2017},
ShuffleNet \citep{zhang2018b},
EffNet \citep{freeman2018}
or DCTI \citep{truong2018},
have at least two orders of magnitude higher number of parameters,
thus amplifying issues of
computational complexity,
model structure,
weight symmetry,
prior specification,
posterior shape,
MCMC convergence
and
sampling effectiveness.

\subsubsection{Model definition.}

An MLP is a feedforward neural network consisting of an input layer, one or
more hidden layers and an output layer
\citep{rosenblatt1958,minsky1988,hastie2016}.
Let $\rho \ge 2$ be a natural number.
Consider an index $j\in\{0,1,\dots,\rho\}$ indicating the layer,
where $j=0$ refers to the input layer, $j=1,2,\dots,\rho-1$ to one of the
$\rho-1$ hidden layers
and $j=\rho$ to the output layer.
Let $\kappa_{j}$ be the number of neurons in layer $j$
and use $\kappa_{0:\rho} = (\kappa_{0},\kappa_{1},\dots,\kappa_{\rho})$
as a shorthand for the sequence of neuron counts per layer.
Under such notation,
$\mbox{MLP}(\kappa_{0:\rho})$
refers to an MLP with $\rho-1$ hidden layers and $\kappa_j$ neurons at layer
$j$.

An $\mbox{MLP}(\kappa_{0:\rho})$
with $\rho-1\ge 1$ hidden layers and
$\kappa_j$ neurons at layer $j$
% $j\in\{0,1,\dots,\rho\}$
is defined recursively as
\begin{align}
\label{mlp_g}
g_{j}(x_{i},\theta_{1:j}) &=
W_{j}h_{j-1}(x_{i},\theta_{1:j-1})+b_{j},\\
\label{mlp_h}
h_{j}(x_{i}, \theta_{1:j}) &=
\phi_{j}(g_{j}(x_{i},\theta_{1:j})),
\end{align}
for $j=1,2,\dots,\rho$. A data point $x_{i}\in\mathbb{R}^{\kappa_0}$
corresponds to the input layer $h_{0}(x_{i})=x_{i}$,
yielding the sequence $g_{1}(x_{i}, \theta_{1})=W_{1}x_{i}+b_{1}$
in the first hidden layer.
$W_{j}$ and $b_{j}$ are the respective weights and biases at layer
$j=1,2,\dots,\rho,$
which constitute the parameters $\theta_{j} = (W_{j}, b_{j})$
at layer $j$.
The shorthand
$\theta_{1:j} = (\theta_{1},\theta_{2},\dots,\theta_{j})$ denotes
all weights and biases up to layer $j$.
Functions
% $\phi_{j}(g_{j})$,
$\phi_{j}$,
known as \textit{activations}, are applied
elementwise to their input $g_{j}$.

The default recommendation of activation in neural networks is a rectified
linear unit (ReLU),
see for instance \citet{jarrett2009,nair2009,goodfellow2016}.
Other activations are the ELU, leaky RELU, tanh and sigmoid
\citep{nwankpa2018}.
If an activation is not present at layer $j$,
then
the identity function $\phi_{j}(g_{j})=g_{j}$
is used as $\phi_{j}$ in \eqref{mlp_h}.

The weight matrix $W_{j}$ in \eqref{mlp_g} has $\kappa_{j}$ rows and
$\kappa_{j-1}$ columns,
while the vector $b_{j}$ of biases has length $\kappa_{j}$.
Concatenating all $\theta_{j}$ across hidden and output layers gives a
parameter vector $\theta=\theta_{1:\rho}\in\mathbb{R}^n$
of length $n=\sum_{j=1}^{\rho}\kappa_{j}(\kappa_{j-1}+1)$.
To define $\theta$ uniquely,
the convention to traverse weight matrix elements row-wise is made.
Apparently, each of $g_{j}$ in \eqref{mlp_g} and $h_{j}$ in \eqref{mlp_h}
has length $\kappa_{j}$.

The notation $W_{j,k,l}$ is introduced to point to the
$(k,l)$-the element of weight matrix $W_{j}$ at layer $j$.
Analogously, $b_{j,k}$ points to the $k$-th coordinate
of bias vector $b_{j}$ at layer $j$.

% Stopped here

\subsubsection{Likelihood for binary classification.}

Consider $s$ samples $(x_{i}, y_{i}),~i=1,2,\dots,s,$
consisting of some input $x_{i}\in\mathbb{R}^{\kappa_0}$ and of a binary
output $y_{i}\in\{0,1\}$.
An $\mbox{MLP}(\kappa_0,\kappa_1,\dots,\kappa_{\rho}=1)$ with a single neuron
in its output layer
can be used for setting the likelihood function
$L(y_{1:s}|x_{1:s},\theta)$
of labels $y_{1:s}=(y_{1},y_{2},\dots,y_{s})$ given the input
$x_{1:s}=(x_{1},x_{2},\dots,x_{s})$ and MLP parameters $\theta$.

Firstly, the \textit{sigmoid activation function}
$\phi_{\rho}(g_{\rho})=1 / (1+\exp{(-g_{\rho})})$
is applied at the output layer of the MLP.
So, the \textit{event probabilities} $\mbox{Pr}(y_{i}=1 | x_{i},\theta)$
are set to
\begin{equation}
\label{bc_mlp_probs}
\begin{split}
\mbox{Pr}(y_{i}=1 | x_{i},\theta)
& = h_{\rho}(x_{i},\theta)
=\phi_{\rho}(g_{\rho}(x_{i},\theta))\\
& =\frac{1}{1+\exp{(-g^{(\rho)}(x^{(i)} , \theta))}}.
\end{split}
\end{equation}

Assuming that the labels are
outcomes of $s$ independent draws from Bernoulli
probability mass functions
with event probabilities given by \eqref{bc_mlp_probs},
the likelihood becomes
%\begin{equation}
%\label{bc_mlp_lik}
%\begin{split}
%L(y_{1:s} | x_{1:s},\theta) &=
%% \prod_{i=1}^{s}f_i(y_i|a_i(x_i,\theta))=
%\prod_{i=1}^{s}\Big( (h_{\rho}(x_{i},\theta))^{y_{i}} \\ % \right. \\
%% & \left.
%& (1-h_{\rho}(x_{i},\theta))^{1-y_{i}} \Big).
%\end{split}
%\end{equation}
\begin{equation}
\label{bc_mlp_lik}
L(y_{1:s} | x_{1:s},\theta) =
\prod_{i=1}^{s}\prod_{k=1}^{2}
(z_{\rho,k}(x_{i},\theta))^{\mathbbm{1}_{\{y_{i}=k-1\}}}.
% \prod_{i=1}^{s}f_i(y_i|a_i(x_i,\theta))=
%\prod_{i=1}^{s}h_{\rho}^{y_{i}}(x_{i},\theta)
%% & \left.
% (1-h_{\rho}(x_{i},\theta))^{1-y_{i}}.
\end{equation}
$z_{\rho,k}(x_{i},\theta),~k=1,2,$
denotes the $k$-th coordinate of
the vector
$z_{\rho}(x_{i},\theta)=
(1-h_{\rho}(x_{i},\theta),h_{\rho}(x_{i},\theta))$
of event probabilities for sample $i=1,2,\dots,s$.
Furthermore,
$\mathbbm{1}$ denotes the indicator function, that is
$\mathbbm{1}_{\{y_{i}=k-1\}}=1$ if $y_{i}=k-1$,
and
$\mathbbm{1}_{\{y_{i}=k-1\}}=0$
otherwise.
The log-likelihood
%$
%\ell(\{y^{(i)}\}|\{x^{(i)}\},\theta):=\log{(L(\{y^{(i)}\}|\{x^{(i)}\},\theta))}
%$
follows as
\begin{equation}
\label{bc_mlp_loglik}
\ell(y_{1:s}|x_{1:s},\theta)
=
%\sum_{i=1}^{s}y^{(i)}\log{(h^{(\rho)}(x^{(i)},\theta))}+
%(1-y^{(i)})\log{(1-h^{(\rho)}(x^{(i)},\theta))}.
\sum_{\substack{i=1 \\ k=1}}^{\substack{s \\2 }}
\mathbbm{1}_{\{y_{i}=k-1\}}
\log{(z_{\rho,k}(x_{i},\theta))}.
\end{equation}

The negative value of log-likelihood \eqref{bc_mlp_loglik}
is known as the \textit{binary cross entropy} (BCE).
To infer the parameters $\theta$ of
$\mbox{MLP}(\kappa_0,\kappa_1,\dots,\kappa_{\rho}=1)$,
the binary cross entropy or a different loss function is minimized using
stochastic optimization methods,
such as stochastic gradient descent (SGD).

\subsubsection{Likelihood for multiclass classification.}

Let $y_i\in\{1,2,\dots,\kappa_{\rho}\}$ be an output variable, which can
take $\kappa_{\rho}\ge 2$ values.
Moreover, consider an $\mbox{MLP}(\kappa_{0:\rho})$ with $\kappa_{\rho}$
neurons in its output layer.

Initially, a \textit{softmax activation function}
$\phi_{\rho}(g_{\rho})=\exp{(g_{\rho})} /
\sum_{k=1}^{\kappa_{\rho}}\exp{(g_{\rho,k})}$
is applied at the output layer of the MLP,
where $g_{\rho,j}$ denotes the $k$-th coordinate of the
$\kappa_{\rho}$-length vector $g_{\rho}$.
Thus, the event probabilities
$\mbox{Pr}(y_{i}=k|x_{i},\theta)$ are
\begin{equation}
\label{mc_mlp_probs}
\begin{split}
\mbox{Pr}(y_{i}=k|x_{i},\theta)
& = h_{\rho,k}(x_{i},\theta)\\
& = \phi_{\rho}(g_{\rho,k}(x_{i},\theta))\\
& =
\frac{\exp{(g_{\rho,k}(x^{(i)} ,
\theta))}}
{\sum_{r=1}^{\kappa_{\rho}}\exp{(g_{\rho,r}(x_{i}, \theta))}}.
\end{split}
\end{equation}
$ h_{\rho,k}(x_{i},\theta)$ denotes
the $k$-th coordinate of the MLP output $h_{\rho}(x_{i},\theta)$.

It is assumed that the labels are outcomes of $s$ independent draws from
categorical probability mass functions
with event probabilities given by \eqref{mc_mlp_probs}, so
the likelihood is
\begin{equation}
\label{mc_mlp_lik}
L(y_{1:s}|x_{1:s},\theta)=
\prod_{i=1}^{s}\prod_{k=1}^{\kappa_{\rho}}
(h_{\rho,k}(x_{i},\theta))^{\mathbbm{1}_{\{y_{i}=k\}}}.
\end{equation}
%$\mathbbm{1}$ denotes the indicator function, that is
%$\mathbbm{1}(y^{(i)}=k)=1$ if $y^{(i)}=k$, and $\mathbbm{1}(y^{(i)}=k)=0$
%otherwise.
The log-likelihood follows as
\begin{equation}
\label{mc_mlp_loglik}
\ell(y_{1:s}|x_{1:s},\theta)
=
%\sum_{i=1}^{s}
%\sum_{k=1}^{\kappa_{\rho}}
\sum_{\substack{i=1 \\ k=1}}^{\substack{s \\\kappa_{\rho}}}
\mathbbm{1}_{\{y_{i}=k\}}
\log{(h_{\rho,k}(x_{i},\theta))}.
\end{equation}
The negative value of log-likelihood \eqref{mc_mlp_loglik} is known as
\textit{cross entropy},
and it is used as loss function for stochastic optimization in multiclass
classification MLPs.

%It is noted that for $\kappa_{\rho}=2$,
An $\mbox{MLP}(\kappa_0,\kappa_1,\dots,\kappa_{\rho}=2)$
with two neurons at the output layer,
event probabilities given by softmax activation \eqref{mc_mlp_probs}
and log-likelihood \eqref{mc_mlp_loglik}
can be used for binary classification.
Such a formulation is an alternative to an
$\mbox{MLP}(\kappa_0,\kappa_1,\dots,\kappa_{\rho}=1)$
with one neuron at the output layer,
event probabilities given by sigmoid activation \eqref{bc_mlp_probs}
and log-likelihood \eqref{bc_mlp_loglik}.
The difference between the two MLP models is the parameterization of event
probabilities,
since a categorical distribution
with $\kappa_{\rho}=2$ levels otherwise coincides with a Bernoulli distribution.

% Sections \ref{mcmc_algorithms} and \ref{mcmc_diagnostics}
% describe the respective MCMC algorithms and MCMC diagnostics
% used in the examples of this paper.

\subsection{MCMC sampling for parameter estimation}
\label{mcmc_algorithms}

Interest is in sampling from the
% unnormalized
parameter posterior
$p(\theta | x_{1:s}, y_{1:s})
\propto
L(y_{1:s} | x_{1:s}, \theta)
\pi (\theta)
$
of a neural network
given the neural network likelihood
$L(y_{1:s} | x_{1:s}, \theta)$
and parameter prior $\pi(\theta)$.
For MLPs, the likelihood $L(y_{1:s} | x_{1:s}, \theta)$
for binary and multiclass classification is provided by
\eqref{bc_mlp_lik} and \eqref{mc_mlp_lik}, respectively.

The parameter posterior $p(\theta | x_{1:s}, y_{1:s})$ is alternatively
denoted by $p(\theta | D_{1:s})$ for brevity.
$D_{1:s}=(x_{1:s}, y_{1:s})$ is a dataset of size $s$ consisting of
input $x_{1:s}$ and output $y_{1:s}$.

%Both examples in this paper rely on an MLP model,
%one consisting of nine and one consisting of twenty seven parameters.
%MCMC inference has been performed on MLPs before, see for example
%\citet{freitas1999,vehtari2000}.
%Manifold Langevin Monte Carlo methods and power posteriors
%have not been used in the context of MLPs.
%The use of such contemporary geometric and population MCMC methods in neural
%networks
%is not an end in itself,
%it is a means for acquiring an understanding of the challenges
%arising from the application of MCMC methods on neural networks
%and for developing benchmark tools for more scalable MCMC algorithms.
%% Domain experts interested in designing more scalable MCMC algorithms
%% can benefit from being aware of issues arising in MCMC inference for neural
%%networks.

This subsection provides an introduction to
the MCMC algorithms and MCMC diagnostics
used in the examples of section \ref{examples}.
Three MCMC algorithms are outlined,
namely Metropolis-Hastings, Hamiltonian Monte Carlo,
and power posterior sampling.
Two MCMC diagnostics are described,
the multivariate potential scale reduction factor (PSRF)
and the multivariate effective sample size (ESS).

\subsubsection{Metropolis-Hastings algorithm.}

One of the most general algorithms for sampling from a posterior
$p(\theta | D_{1:s})$ is the Metropolis-Hastings (MH) algorithm
\citep{metropolis1953, hastings1970}.
Given the current state $\theta$,
the MH algorithm initially samples a state $\theta^{*}$ from a
\textit{proposal density} $g_{\theta}$
and subsequently accepts the proposed state $\theta^{*}$ with probability
\begin{equation*}
\left\{
\begin{array}{ll}
\min \left\{
\frac{p(\theta^{*}|D_{1:s}) g_{\theta^{*}} (\theta)}
{p(\theta|D_{1:s})g_{\theta} (\theta^{*})}
, 1
\right\}  & \mbox{if } p(\theta|D_{1:s})g_{\theta} (\theta^{*})>0, \\
1 & \mbox{otherwise}.
\end{array}
\right.
\end{equation*}

Typically, a normal proposal density $g_{\theta}=\mathcal{N}(\theta, \Lambda)$
with a constant covariance matrix $\Lambda$ is used.
For such a normal $g_{\theta}$,
the acceptance probability simplifies to
$\min \left\{p(\theta^{*}|D_{1:s})/p(\theta|D_{1:s}), 1\right\}$,
yielding the so called \textit{random walk Metropolis} algorithm.

\subsubsection{Hamiltonian Monte Carlo.}

Hamiltonian Monte Carlo (HMC) draws samples
from an augmented parameter space
via Gibbs steps,
by computing a trajectory in the parameter space
according to Hamiltonian dynamics.
For a more detailed review of HMC, see \citet{neal2011}.

\subsubsection{Power posterior sampling.}

Power posterior (PP) sampling by \citet{friel2008} is a population Monte Carlo algorithm.
It involves $m+1$ chains
drawn from tempered versions $p^{t_i}(\theta |D_{1:s})$
of a target posterior $p(\theta |D_{1:s})$
for a \textit{temperature schedule} $t_i\in[0,1],~i\in\{0,1,\dots,m\}$, where $t_m=1$.
At each iteration, the state of each chain is updated using an MCMC sampler associated with that chain
and subsequently states between pairs of chains are swapped according to an MH algorithm.
For the $i$-th chain, a sample $j$ is drawn from a probability mass function $p_i$
with probability $p_i(j)$, in order to determine
the pair $(i, j)$ for a possible swap.

\textit{Power posteriors} $p^{t_i}(\theta|D_{1:s}),~t_i < t_m,$ are smooth approximations of
the target density $p^{t_m}(\theta |D_{1:s})=p(\theta |D_{1:s})$,
facilitating exploration of the parameter space via state transitions
between chains of $p^{t_i}(\theta|D_{1:s})$ and of $p(\theta |D_{1:s})$.
In this paper,
a categorical probability mass function $p_i$ is used in PP sampling
for determining candidate pairs of chains for state swaps
(see \ref{appendix_categorical}).

\subsubsection{Multivariate PSRF.}
\label{mcmc_diagnostics}

PSRF, commonly denoted by $\hat{R}$,
is an MCMC diagnostic of convergence conceived by \citet{gelman1992} and extended to its multivariate version by \citet{brooks1998}.
This paper uses the multivariate PSRF by \citet{brooks1998},
% The multivariate PSRF by \citet{brooks1998}
which provides a single-number summary of convergence
across the $n$ dimensions of a parameter,
requiring a Monte Carlo covariance matrix estimator for the parameter.
%The idea behind PSRF is that the variance of all the chains will be higher than the variance of individual chains,
%if convergence has not been reached.

%Univariate variants of PSRF
%\citep{gelman1992, stan2019, vehtari2019}
%are computed
%for each scalar coordinate of an $n$-dimensional parameter,
%requiring a Monte Carlo variance estimator for each coordinate.

To acquire the multivariate PSRF,
the \textit{multivariate initial monotone sequence estimator} (MINSE)
of Monte Carlo covariance is employed
\citep{dai2017}.
%The MINSE is a multivariate extension of the
%\textit{initial monotone sequence estimator}
%of Monte Carlo variance \citep{geyer1992}.
In a Bayesian setting,
the MINSE estimates
the covariance matrix of a parameter posterior
$p(\theta | D_{1:s})$.

%A short description of multivariate PSRF follows;
%for more details see \citet{brooks1998}.
%Let
%% $\omega_{j,1:v}\in\mathbb{R}^{n\times v},~j=1,2,\dots,m,$
%$\omega_{j,1:v},~j=1,2,\dots,m,$
%be the $j$-th Markov chain realization of length $v$
%for a parameter $\theta\in\mathbb{R}^n$.
%% among $m$ such realizations.
%Furthermore, let
%$\bar{\omega}_{j,.}=\sum_{k=1}^{v}\omega_{j,k}/v$
%be the mean of chain $j$ and
%$\bar{\omega}_{.,.}=\sum_{j=1}^{m}\bar{\omega}_{j,.}/m$
%the mean of means across chains.
%Within-chain covariance is captured by the mean
%$A=\sum_{j=1}^{m}C_j$, where $C_j$ is the MINSE for chain $j$.
%% $A=\sum_{j=1}^{m}C_j$ of INSE estimators $C_j,~j=1,2,\dots,m,$
%% across the $m$ chains.
%Between-chain covariance is captured by
%the empirical covariance matrix
%\begin{equation*}
%B=\frac{1}{m-1}
%\sum_{j=1}^{m}
%(\bar{\omega}_{1:n,j,.}-\bar{\omega}_{1:n,.,.})
%(\bar{\omega}_{1:n,j,.}-\bar{\omega}_{1:n,.,.})
%^{\prime}.
%\end{equation*}
%The multivariate PSRF is given by
%\begin{equation}
%\label{eq_psrf}
%\hat{R}=
%\frac{v-1}{v}
%+\left(\frac{m+1}{m}\right)\lambda_{1},
%\end{equation}
%where $\lambda_1$ is the largest eigenvalue of $A^{-1}B$.

To compute PSRF, several independent Markov chains are simulated.
\citet{gelman2004} recommend terminating MCMC sampling as soon as $\hat{R}<1.1$.
More recently, \citet{vats2018a} make an argument based on ESS that a cut-off of $1.1$ for $\hat{R}$ is too high
to estimate a Monte Carlo mean with reasonable uncertainty.
\citet{vehtari2019} recommend simulating at least $m=4$ chains to compute $\hat{R}$
and using a threshold of $\hat{R} < 1.01$.
% In this paper,
% split-$\hat{R}$ and folded-split-$\hat{R}$ are reported.
% employing a threshold of $1.01$.

% \subsubsection{Effective sample size}
\subsubsection{Multivariate ESS.}

The ESS of an estimate obtained from a Markov chain realization
is interpreted as the number of independent samples
that provide an estimate with variance equal
to the variance of the estimate obtained from the Markov chain realization.
For a more extensive treatment entailing univariate
approaches to ESS, see
\citet{vats2018b,gong2016,kass1998}.

$\hat{R}$ and its variants can fail to diagnose poor mixing of a Markov chain,
whereas low values of ESS are an indicator of poor mixing.
It is thus recommended to check both $\hat{R}$ and ESS \citep{vehtari2019}.
For a theoretical treatment of the relation between $\hat{R}$ and ESS, see \citet{vats2018a}.

Univariate ESS pertains to a single coordinate of an $n$-dimensional parameter.
\citet{vats2019} introduce a multivariate version of ESS,
which provides a single-number summary of sampling effectiveness
across the $n$ dimensions of a parameter.
Similarly to multivariate PSRF \citep{brooks1998},
multivariate ESS \citep{vats2019}
requires a Monte Carlo covariance matrix estimator
for the parameter.

Given a single Markov chain realization of length $v$
for an $n$-dimensional parameter,
\citet{vats2019} define multivariate ESS as
\begin{equation*}
% \label{eq_ess}
\hat{S}=v
\left(
\frac{\det{(E)}}{\det{(C)}}
\right)^{1/n}.
\end{equation*}
$\det{(E)}$ is the determinant of
the empirical covariance matrix $E$ and
$\det{(C)}$ is the determinant of
a Monte Carlo covariance matrix estimate $C$
for the chain.
In this paper,
the multivariate ESS by \citet{vats2019} is used,
setting $C$ to be the MINSE for the chain.

\subsection{Bayesian marginalization for prediction}
\label{subsec_bm}

This subsection briefly reviews
the notion of posterior predictive distribution
based on Bayesian marginalization,
posterior predictive distribution approximation via Monte Carlo integration,
and associated binary and multiclass classification.

\subsubsection{Posterior predictive distribution.}

Consider a set
$D_{1:s}=(x_{1:s}, y_{1:s})$ of $s$
training data points
and a single test data point $(x, y)$
consisting of some test input $x$
and test output $y$.
Integrating out the parameters $\theta$
of a model fitted to $D_{1:s}$ yields
the posterior predictive distribution
% $p(y|x,D_{1:s})$ of test output $y$ given $(x, D_{1:s}$
\begin{equation}
\label{pred_posterior}
% p(y|x,D_{1:s})=
\underbrace{p(y|x,D_{1:s})}_\text{
\parbox{2cm}{\centering Predictive\\[-4pt]distribution}} =
\int
\underbrace{p(y | x, \theta)}_\text{Likelihood}
\underbrace{p(\theta | D_{1:s})}_\text{
\parbox{2cm}{\centering Parameter\\[-4pt]posterior}}
d\theta .
\end{equation}
% In \eqref{pred_posterior},
% the posterior predictive distribution $p(y|x,D_{1:s})$
% is expressed as an integral of the product between
% likelihood $p(y | x, \theta)$
% and parameter posterior $p(\theta | D_{1:s})$
% with respect to model parameters $\theta$.
\ref{appendix_predictive} provides a derivation
of \eqref{pred_posterior}.

\subsubsection{Monte Carlo approximation.}

\eqref{pred_posterior} can be written as
\begin{equation}
\label{pred_posterior_expectation}
p(y|x,D_{1:s})=
\mbox{E}_{\theta | D_{1:s}} [p(y | x, \theta)] .
\end{equation}
\eqref{pred_posterior_expectation} states the
posterior predictive distribution $p(y|x,D_{1:s})$
as an expectation
of the likelihood $p(y | x, \theta)$
evaluated at the test output $y$
with respect to the parameter posterior
$p(\theta | D_{1:s})$ learnt from
the training set $D_{1:s}$.

The expectation in \eqref{pred_posterior_expectation}
can be approximated via Monte Carlo integration.
More specifically, a Monte Carlo approximation
of the posterior predictive distribution is given by
\begin{equation}
\label{pred_posterior_approx}
p(y|x,D_{1:s}) \simeq
\sum_{k=1}^{v} p(y | x, \omega_{k}) .
\end{equation}
The sum in \eqref{pred_posterior_approx}
involves evaluations of
the likelihood across $v$ iterations
$\omega_k,~k=1,2,\dots,v,$
of a Markov chain realization
$\omega_{1:v}$
obtained from the parameter posterior
$p ({\theta | D_{1:s}})$.

\subsubsection{Classification rule.}

In the case of binary classification, the prediction
$\hat{y}$ for the test label $y\in\{0,1\}$
is
\begin{equation}
\label{bin_class_pred}
\hat{y} =
\left\{
\begin{array}{ll}
1  & \mbox{if } p(y|x,D_{1:s}) \geq 0.5, \\
0 & \mbox{otherwise}.
\end{array}
\right.
\end{equation}
For multiclass classification,
the prediction label
$\hat{y}$ for the test label $y\in\{1,2,\dots,\kappa_{\rho}\}$
is
\begin{equation}
\label{multi_class_pred}
\hat{y} =
\argmax_{y} {
\{p(y|x,D_{1:s})\}
}.
\end{equation}

The classification rules
\eqref{bin_class_pred} and \eqref{multi_class_pred}
for binary and multiclass classification
maximize the posterior predictive distribution.
This way, predictions are made based on the Bayesian principle.
The uncertainty of predictions is quantified,
since the posterior predictive probability $p(y|x,D_{1:s})$
of each predicted label $\hat{y}$ is available.

\section{Examples}
\label{examples}

%Two examples of MLPs are used for showcasing challenges in MCMC inference for neural networks.
%An $\mbox{MLP}(2, 2, 1)$ and an $\mbox{MLP}(4, 3, 3)$ are used
%in the context of a binary and of a multiclass classification example, respectively.
%The focus
%% of the examples, and of the paper more generally,
%of this paper is on inferring the parameter posterior
%$p(\theta | \{x^{(i)}\}, \{y^{(i)}\})$
%of an MLP using MCMC,
%rather than inferring the predictive posterior.

Four examples of Bayesian inference for MLPs based on MCMC
are presented.
A different dataset is used for each example.
The four datasets entail
simulated noisy data from the exclusive-or (XOR) function,
and observations collected from Pima Indians, penguins and hawks.
Section \ref{mlp_datasets} introduces the four datasets.
Each of the four datasets is split into a training and a test set
for parameter inference and for predictions, respectively.
MLPs with one neuron in the output layer are fitted to
the noisy XOR and Pima datasets
to perform binary classification,
while MLPs with three neurons in the output layer are fitted
to the penguin and hawk datasets
to perform multiclass classification
with three classes.
Table \ref{data_models_table}
shows the training and test sample sizes of the four datasets,
and the fitted MLP models with their associated number $n$ of parameters.

\begin{table}[t]
\centering
\caption{Training and test sample sizes
of the four datasets of section \ref{examples},
architectures of fitted MLP models and associated number
$n$ of MLP parameters.}
\label{data_models_table}
\begin{tabular}{|l|r|r|l|r|}
 \hline
 \multicolumn{1}{|c|}{\multirow{2}{*}{Dataset}} &
 \multicolumn{2}{c|}{Sample size} & % \cline{2-3}
 \multicolumn{1}{c|}{\multirow{2}{*}{Model}} &
 \multicolumn{1}{c|}{\multirow{2}{*}{$n$}}\\
 \cline{2-3}
 &
 \multicolumn{1}{c|}{Training} &
 \multicolumn{1}{c|}{Test}
 &
 &
 \\
 \hline
 Noisy XOR & $500$ & $120$ & $\mbox{MLP}(2,2,1)$ & $9$ \\ \hline
 Pima & $262$ & $130$ & $\mbox{MLP}(8,2,2,1)$ & $27$ \\ \hline
 Penguins & $223$ & $110$ & $\mbox{MLP}(6,2,2,3)$ & $29$ \\ \hline
 Hawks & $596$ & $295$ & $\mbox{MLP}(6,2,2,3)$ & $29$ \\ \hline
\end{tabular}
%\caption{Caption.}
%\label{num_summaries}
\end{table}

In the examples, samples are drawn via MCMC from the
unnormalized log-posterior
\begin{equation*}
\log{(p(\theta | x_{1:s}, y_{1:s}))}=
\ell (y_{1:s} | x_{1:s}, \theta)
+\log{(\pi(\theta))}
\end{equation*}
of MLP parameters.
%$\ell (\{y^{(i)}\}|\{x^{(i)}\},\theta) + \log{\pi (\theta)}$,
The log-likelihood $\ell (y_{1:s} | x_{1:s}, \theta)$
for binary or multiclass classification corresponds to
\eqref{bc_mlp_loglik} or \eqref{mc_mlp_loglik}.
$\log{(\pi(\theta))}$ is the log-prior of MLP parameters.

% \subsection{MLP for exclusive-or data}

\subsection{Datasets}
\label{mlp_datasets}

An introduction to the four datasets used in this paper follows.
The simulated noisy XOR dataset does not contain missing values,
while the real datasets for Pima, penguins and hawks come with missing values.
%Input variables (\textit{features})
%with relatively large number of missing values
%have been removed from the three real datasets to
%avoid dropping large number of data points.
Data points containing missing values in the chosen variables
have been dropped from the three real datasets.
All \textit{features} (input variables) in the three real datasets
have been standardized.
The four datasets,
in their final form used for inference and prediction,
are available at
\url{https://github.com/papamarkou/bnn_mcmc_examples}.

\subsubsection{XOR dataset.}

% $\mbox{MLP}(2, 2, 1)$. Sigmoid activation after hidden layer. Number of Monte Carlo iterations.

The so called XOR function
$f:\{0,1\}\times \{0,1\}\rightarrow \{0,1\}$
returns $1$ if exactly one of its binary input values is equal to $1$,
otherwise it returns $0$.
The $s=4$ data points defining XOR are
$(x_{1}, y_{1})=((0,0), 0)$,
$(x_{2}, y_{2})=((0,1), 1)$,
$(x_{3}, y_{3})=((1,0), 1)$ and
$(x_{4}, y_{4})=((1,1), 0)$.

A perceptron without a hidden layer can not learn the XOR function \citep{minsky1988}.
On the other hand, an $\mbox{MLP}(2, 2, 1)$ with a single hidden layer of two neurons
can learn the XOR function \citep{goodfellow2016}.

An $\mbox{MLP}(2, 2, 1)$ has a parameter vector $\theta$ of length $n=9$,
as $W_{1},b_{1},W_{2}$ and $b_{2}$
have respective dimensions $2\cdot 2, 2\cdot 1, 2\cdot 1$ and $1\cdot 1$.
Since the number $s=4$ of data points defined by the exact XOR function
is less than the number $n=9$ of parameters in the fitted $\mbox{MLP}(2, 2, 1)$,
the parameters can not be fully identified.

To circumvent the lack of identifiability
arising from the limited number of data points,
a larger dataset is simulated by introducing a noisy version of XOR.
% One way of simulating noisy XOR data points is
% by introducing an auxiliary function
Firstly, consider the auxiliary function
$\psi : [-c, 1+c]\times [-c, 1+c]\rightarrow \{0,1\}\times \{0,1\}$
given by
\begin{align*}
\psi(u-c,u-c) &= (0, 0),\\
\psi(u-c,u+c) &= (0, 1),\\
\psi(u+c,u-c) &= (1, 0),\\
\psi(u+c,u+c) &= (1, 1).
\end{align*}
$\psi$ is presented in parametrized form, in terms of
a constant $c\in (0.5, 1)$ and
a uniformly distributed random variable $u\sim\mathcal{U}(0,1)$.
The noisy XOR function is then defined as the function composition $f\circ\psi$.

A training and a test set of noisy XOR points,
generated using $f\circ\psi$ and $c=0.55$,
are shown in figure \ref{noisy_xor_scatterplots}.
$125$ and $30$ noisy XOR points
per exact XOR point $(x_i,y_i),~i=1,2,3,4,$
are contained in the training and test set, respectively.
So, the training and test sample sizes
are $500$ and $120$,
as reported in table \ref{data_models_table} and
as visualized in figure \ref{noisy_xor_scatterplots}.

In figure \ref{noisy_xor_scatterplots},
the training and test sets of noisy XOR points
consist of two input variables
$(u\pm 0.55, u\pm 0.55)\in [-0.55, 1.55]\times [-0.55, 1.55]$
and of one output variable $f\circ\psi (u\pm 0.55, u\pm 0.55)\in\{0,1\}$.
The four colours classify noisy XOR input $(u\pm 0.55, u\pm 0.55)$
with respect to the corresponding exact XOR input
$\psi(u\pm 0.55, u\pm 0.55)\in\{(0,0),(0,1),(1,0),(1,1)\}$;
the two different shapes classify noisy XOR output,
with circle and triangle corresponding to $0$ and $1$.

%MCMC is run to learn the posterior of $\theta$ given the four XOR data points %$\{(x^{(i)},y^{(i)}):i=1,2,3,4\}$.
%The sigmoid function is used as activation $\phi^{(1)}$ on the hidden layer,
%since it achieves higher acceptance rate in the XOR example than a ReLU,
%according to MCMC pilot runs.

% Interest is in assessing the capacity of MCMC to learn the parameters $\mbox{MLP}(2, 2, 1)$
% rather than predicting the output of XOR.

\subsubsection{Pima dataset.}

The Pima dataset contains observations
taken from female patients of Pima Indian heritage.
The binary output variable indicates whether or not a patient has diabetes.
Eight features are used as diagnostics of diabetes,
namely
the number of pregnancies,
plasma glucose concentration,
diastolic blood pressure,
triceps skinfold thickness,
insulin level,
body mass index,
diabetes pedigree function and
age.

For more information about the Pima dataset, see
\citet{smith1988}.
The original data,
prior to removal of missing values and feature standardization,
are available as the \texttt{PimaIndiansDiabetes2} data frame
of the \texttt{mlbench} \texttt{R} package.

\subsubsection{Penguin dataset.}

The penguin dataset consists of body measurements
for three penguin species
observed on three islands in the Palmer Archipelago, Antarctica.
Ad\'{e}lie, Chinstrap and Gentoo penguins
are the three observed species.
Four body measurements per penguin are taken,
specifically body mass, flipper length, bill length and bill depth.
The four body measurements, sex and location (island)
make up a total of six features utilized for deducing
the species to which a penguin belongs.
Thus, the penguin species
is used as output variable.

\citet{horst2020} provide more details about the penguin dataset.
In their original form, prior to data filtering,
the data are available at
\url{https://github.com/allisonhorst/palmerpenguins}.

\subsubsection{Hawk dataset.}

The hawk dataset is composed of observations for three hawk species
collected from
Lake MacBride near Iowa City, Iowa.
Cooper's, red-tailed and sharp-shinned hawks
are the three observed species.
Age,
wing length,
body weight,
culmen length,
hallux length and
tail length
are the six hawk features employed in this paper
for deducing the species to which a hawk belongs.
So, the hawk species is used as output variable.

\citet{cannon2019} mention that
Emeritus Professor Bob Black at Cornell College
shared the hawk dataset publicly.
The original data, prior to data filtering,
are available as the \texttt{Hawks} data frame
of the \texttt{Stat2Data} \texttt{R} package.

\subsection{Experimental configuration}
\label{subsec_experim_conf}

To fully specify the MLP models of table
\ref{data_models_table}, their activations are listed.
A sigmoid activation function is applied at each hidden layer
of each MLP.
Additionally, a sigmoid activation function is applied
at the output layer of
$\mbox{MLP}(2, 2, 1)$ and
of $\mbox{MLP}(8, 2, 2, 1)$,
conforming to log-likelihood
\eqref{bc_mlp_loglik}
for binary classification.
A softmax activation function is applied
at the output layer of
$\mbox{MLP}(6, 2, 2, 3)$,
in accordance with log-likelihood
\eqref{mc_mlp_loglik}
for multiclass classification.
The same $\mbox{MLP}(6, 2, 2, 3)$ model
is fitted to the penguin and hawk datasets.

A normal prior
$\pi(\theta)=\mathcal{N}(0, 10 I)$ is adopted
for the parameters $\theta\in\mathbb{R}^n$
of each MLP model shown in
table \ref{data_models_table}.
An isotropic covariance matrix $10I$ assigns
relatively high prior variance, equal to $10$,
to each coordinate of $\theta$,
thus setting empirically a
seemingly non-informative prior.

MH and HMC are run for each of the four examples
of table \ref{data_models_table}.
PP sampling incurs higher computational cost than MH and HMC;
for this reason, PP sampling is run only for noisy XOR.
Ten power posteriors are employed
for PP sampling, and MH is used for within-chain moves.
On the basis of pilot runs, the PP temperature schedule
is set to $t_i=1,~i=0,1,\dots,9$; this implies that
each power posterior is set to be the
parameter posterior and consequently
between-chain moves are made among ten chains
realized from the parameter posterior.
Empirical hyperparameter tuning for MH, HMC and PP is carried out.
The chosen
MH proposal variance,
HMC number of leapfrog steps and
HMC leapfrog step size
for each example can be found in
\url{https://github.com/papamarkou/bnn_mcmc_examples}.

$m=10$ Markov chains are realized for each combination
of training dataset shown in table \ref{data_models_table} and
of MCMC sampler.
$110,000$ iterations are run per chain realization,
$10,000$ of which are discarded as burn-in.
Thereby, $v=100,000$ post-burnin iterations
are retained per chain realization.

MINSE computation, required by multivariate PSRF
% \eqref{eq_psrf}
and multivariate ESS,
% \eqref{eq_ess},
is carried out using $v=100,000$ post-burnin iterations
per realized chain.
The multivariate PSRF
% \eqref{eq_psrf}
for each dataset-sampler setting
is computed across the $m=10$ realized chains for the setting.
On the other hand,
the multivariate ESS
% \eqref{eq_ess}
is computed for each realized chain,
and the mean across $m=10$ ESSs is reported
for each dataset-sampler setting.

Monte Carlo approximations of posterior predictive distributions
are computed according to
\eqref{pred_posterior_approx}
for each data point of each test set.
To reduce the computational cost,
the last $v=10,000$ iterations of each realized chain
are used in
\eqref{pred_posterior_approx}.

Predictions for binary and multiclass classification
are made using
\eqref{bin_class_pred} and \eqref{multi_class_pred}, respectively.
Given a single chain realization from an MCMC sampler,
predictions are made for every point in a test set;
the predictive accuracy is then computed as
the number of correct predictions
over the total number of points in the test set.
Subsequently,
the mean of predictive accuracies
across the $m=10$ chains realized from the sampler
is reported for the test set.
% each model-sampler setting.

\subsection{Numerical summaries}
\label{num_summaries}

Table
\ref{num_summaries_table}
shows numerical summaries for
each set of $m=10$ Markov chains realized by an MCMC sampler
for a dataset-MLP combination of table
\ref{data_models_table}.
Multivariate PSRF
% \eqref{eq_psrf}
and multivariate ESS
% \eqref{eq_ess}
diagnose the capacity of MCMC sampling to perform parameter inference.
Predictive accuracy via Bayesian marginalization
\eqref{pred_posterior_approx},
based on
classification rules
\eqref{bin_class_pred}
and
\eqref{multi_class_pred}
for binary and multiclass classification,
demonstrates the predictive performance of MCMC sampling.
The last column of table
\ref{num_summaries_table}
displays the predictive accuracy via
\eqref{pred_posterior_approx}
with samples
$\omega_k,~k=1,2,\dots,v,$ drawn from the prior
$\pi(\theta)=\mathcal{N}(0,10I)$,
thus providing an approximation of
the expected posterior predictive probability
\begin{equation}
\label{pred_posterior_wrt_prior}
\mbox{E}_{\theta}[p(y|x,\theta)]=
\int p(y | x, \theta) \pi(\theta) d\theta
\end{equation}
with respect to prior $\pi(\theta)$.

\begin{table}[t]
\centering
\caption{Multivariate PSRF, multivariate ESS
and predictive accuracy
for each set of ten Markov chains realized by an MCMC sampler
for a dataset-MLP combination.
Predictive accuracies based on samples from the prior
are reported as model-agnostic baselines.
}
\label{num_summaries_table}
\begin{tabular}{|l|r|r|r|r|}
 \hline
 \multicolumn{1}{|c|}{\multirow{2}{*}{Sampler}} &
 \multicolumn{1}{c|}{\multirow{2}{*}{PSRF}} &
 \multicolumn{1}{c|}{\multirow{2}{*}{ESS}} &
 \multicolumn{2}{c|}{Accuracy} \\ \cline{4-5}
 % \midrule
 &
 &
 &
 \multicolumn{1}{c|}{MCMC} &
 \multicolumn{1}{c|}{Prior} \\
 \hline
 \multicolumn{5}{|c|}{Noisy XOR, $\mbox{MLP}(2, 2, 1)$} \\
 \hline
 MH  &  1.2057 &    540 &     75.92 &      \multirow{3}{*}{48.33} \\
 HMC & 13.8689 &  25448 &     74.75 &      \\
 PP  &  2.2885 &   4083 &     87.58 &      \\
 \hline
 \multicolumn{5}{|c|}{Pima, $\mbox{MLP}(8, 2, 2, 1)$} \\
 \hline
 MH  &  1.0007 &     93 &     79.31 &      \multirow{2}{*}{51.69} \\
 HMC &  1.0001 &    718 &     80.38 &      \\
 \hline
 \multicolumn{5}{|c|}{Penguins, $\mbox{MLP}(6, 2, 2, 3)$} \\
 \hline
 MH  &  1.0229 &    217 &    100.00 &      \multirow{2}{*}{36.45} \\
 HMC &  1.6082 &   3127 &    100.00 &      \\
 \hline
 \multicolumn{5}{|c|}{Hawks, $\mbox{MLP}(6, 2, 2, 3)$} \\
 \hline
 MH  &  1.0319 &    168 &     97.97 &      \multirow{2}{*}{28.85} \\
 HMC &  1.4421 &   1838 &     98.03 &      \\
 \hline
\end{tabular}
%\caption{Caption.}
%\label{num_summaries}
\end{table}

PSRF is above $1.01$ \citep{vehtari2019},
indicating lack of convergence,
in three out of four datasets.
% (table \ref{num_summaries_table}).
ESS is low
considering the post-burnin length of
$v=100,000$ of each chain realization,
indicating slow mixing.
MCMC sampling for Pima data
is the only case of attaining PSRF less than $1.01$,
yet the ESS values for Pima are the lowest among
the four datasets.
Overall, simultaneous low PSRF and high ESS are not
reached in any of the examples.

The predictive accuracy is high
in multiclass classification,
despite the lack of convergence and slow mixing.
Bayesian marginalization based on HMC samples
yields $100\%$ and $98.03\%$ predictive accuracy
on the penguin and hawk test datasets,
despite the PSRF values of
$1.6082$ and $1.4421$
on the penguin and hawk training datasets.

PP sampling for the binary classification problem of noisy XOR
leads to higher predictive accuracy ($87.58\%$)
than MH ($75.92\%$) or HMC ($74.75\%$).
The $87.58\%$ predictive accuracy is attained by PP sampling
despite the associated PSRF value of $2.2885$.

Bayesian marginalization based on MCMC sampling
outperforms prior beliefs or random guesses
in terms of predictive inference,
despite MCMC diagnostic failures.
For instance,
Bayesian marginalization
via non-converged HMC chain realizations
yields $74.75\%$, $100\%$ and $98.03\%$ predictive accuracy on
the noisy XOR, penguin and hawk datasets.
Approximating the posterior predictive distribution
with samples from the parameter prior
yields $48.33\%$, $36.45\%$ and $28.85\%$ predictive accuracy on
the same datasets.
It is noted that $48.33\%$ is close to
a $50/50$ random guess for binary classification,
while $36.45\%$ and $28.85\%$ are close to
a $1/3$ random guess for multiclass classification with three classes.

\subsection{Visual summaries for parameters}

Visual summaries for MLP parameters
are presented in this subsection.
In particular, Markov chain traceplots and
a comparison between MCMC sampling and
ensemble training
are displayed.

\subsubsection{Non-converged chain realizations.}

\begin{figure}[t]
	\centering
	\includegraphics[width=1\linewidth]{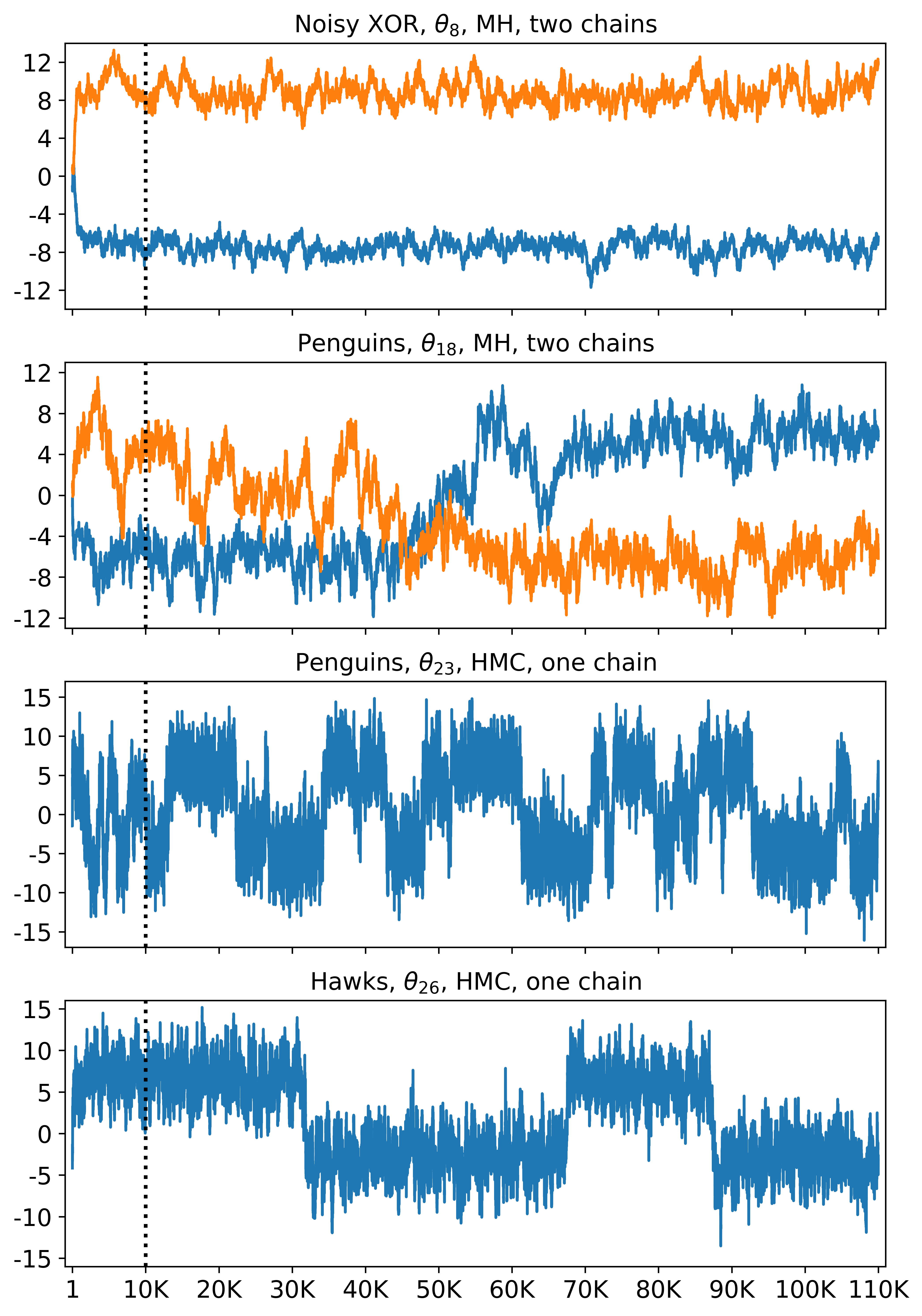}
	\caption{Markov chain traceplots of
    four parameter coordinates
    of MLP models introduced in table
    \ref{data_models_table}.
    The vertical dotted lines indicate the end of burnin.}
	\label{traceplots}
\end{figure}

Figure \ref{traceplots} shows chain traceplots of
four parameters
of MLP models introduced in table
\ref{data_models_table}.
These traceplots visually demonstrate
entrapment in local modes,
mode switching
and more generally lack of convergence.

All $110,000$ iterations per realized chain,
which include burnin, are shown in the traceplots
of figure \ref{traceplots}.
The vertical dotted lines delineate
the first $10,000$ burnin iterations.

Two realized MH chains for parameter $\theta_{8}$
of the $\mbox{MLP}(2, 2, 1)$ model fitted
to the noisy XOR training data are plotted.
The traces in orange and in blue gravitate during burnin towards
modes in the vicinity of $8$ and $-8$, respectively,
and then get entrapped for the entire simulation time
in these modes.
Parameter $\theta_{8}$ corresponds to a weight
connecting a neuron in the hidden layer with
the neuron of the output layer of $\mbox{MLP}(2, 2, 1)$.
The two realized chains for $\theta_{8}$ explore
two regions symmetric about zero associated with
symmetries of weight $\theta_{8}$.

Two realized MH chains for parameter $\theta_{18}$
of the $\mbox{MLP}(6, 2, 2, 3)$ model fitted
to the penguin training data are plotted,
one shown in orange and one in blue.
Each of these two traces
initially explore a mode,
transit to a seemingly symmetric mode
about halfway through the simulation time (post-burnin)
and explore the symmetric mode
in the second half of the simulation.

One HMC chain traceplot for parameter $\theta_{23}$
and one HMC chain traceplot for parameter $\theta_{26}$
of the $\mbox{MLP}(6, 2, 2, 3)$ model fitted to the
penguin and hawk training data, respectively, are shown.
The traces of these two parameters exhibit similar behaviour,
each of them switching between two symmetric regions about zero.
% One difference between these two traces is
% the frequency of mode switching occurrences.

Switching between symmetric modes,
as seen in the displayed traceplots,
manifests weight symmetries.
These traceplots exemplify how computational time
is wasted during MCMC to explore
equivariant parameter posterior modes of a neural network
\citep{nalisnick2018}.
Consequently, the realized chains do not converge.

\subsubsection{MCMC sampling vs ensemble training.}

An exemplified comparison between MCMC sampling and
ensemble training for neural networks follows.
To this end, the same noisy XOR training data
and the same $\mbox{MLP}(2, 2, 1)$ model,
previously used for MCMC sampling,
are used for ensemble training.

\begin{figure}[t]
	\begin{subfigure}{.491\textwidth}
		\centering
		\includegraphics[width=1\linewidth]{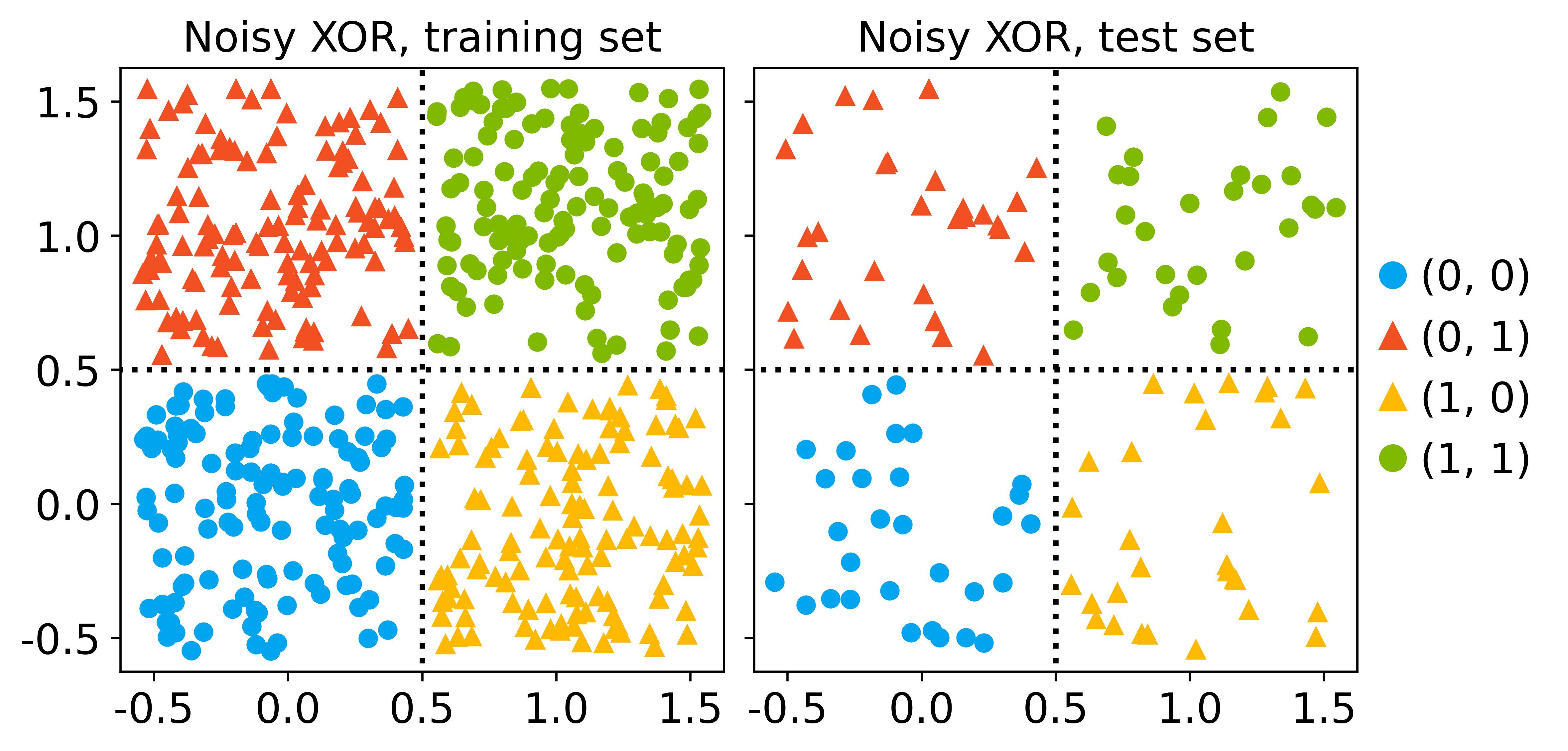}
		\caption{Noisy XOR training set (left) and test set (right).}
		\label{noisy_xor_scatterplots}
	\end{subfigure}\\
	\begin{subfigure}{.491\textwidth}
		\centering
		\includegraphics[width=1\linewidth]{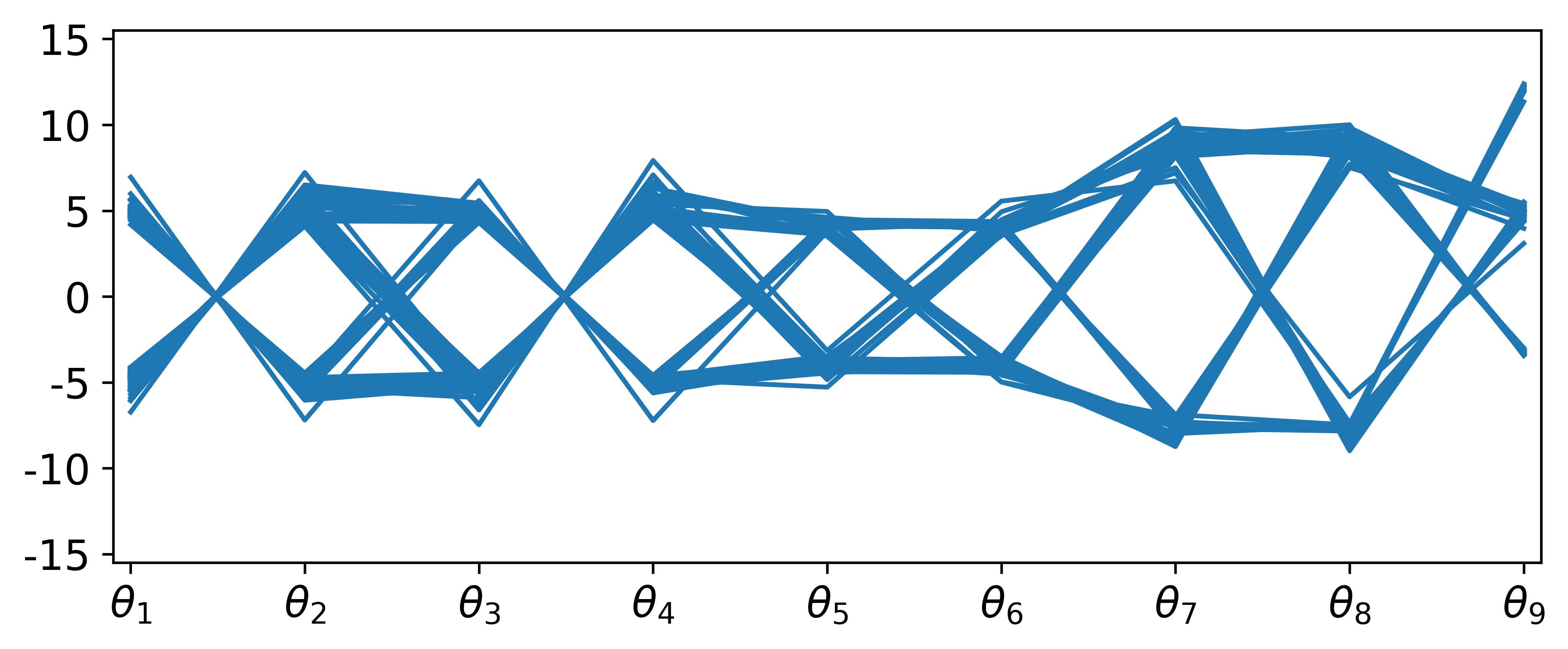}
		\caption{$100$ SGD solutions from training $\mbox{MLP}(2, 2, 1)$.}
		\label{noisy_xor_parallel_coords_plot}
	\end{subfigure}\\
  \begin{subfigure}{.491\textwidth}
  \centering
  \includegraphics[width=1\linewidth]{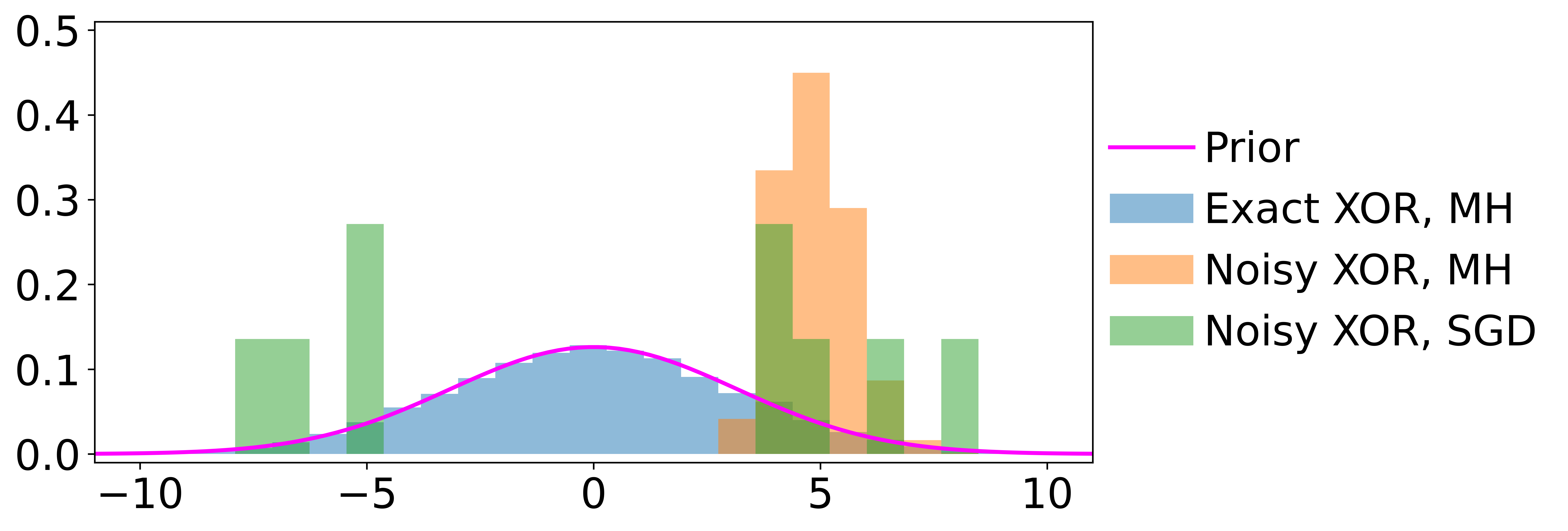}
  \caption{Histograms of parameter $\theta_{3}$ of $\mbox{MLP}(2, 2, 1)$.}
  \label{noisy_xor_marginal_par_posteriors}
  \end{subfigure}
	\caption{Comparison between MH sampling and ensemble training
	of an $\mbox{MLP}(2, 2, 1)$ model fitted to noisy XOR data.
	SGD is used for ensemble training. Each accepted SGD solution
	has predictive accuracy above $85\%$ on the noisy XOR test set.}
	\label{vis_summaries_mcmc_vs_optim}
\end{figure}

To recap, the noisy XOR dataset
is introduced in subsection \ref{mlp_datasets} and
is displayed in figure \ref{noisy_xor_scatterplots};
a sigmoid activation function is applied to the
hidden and output layer of $\mbox{MLP}(2, 2, 1)$,
and the BCE loss function is employed,
which is the negative value of log-likelihood
\eqref{bc_mlp_loglik}.

Ensemble learning is conducted by training
the $\mbox{MLP}(2, 2, 1)$ model
on the noisy XOR training set
multiple times.
At each training session,
SGD is used for minimizing the BCE loss.
SGD is initialized by drawing a sample from
$\pi(\theta)=\mathcal{N}(0, 10I)$,
which is the same density used as prior for MCMC sampling.
$2,000$ epochs are run per training session,
with a batch size of $50$ and a learning rate of $0.002$.
The SGD solution from the training session is accepted if its
predictive accuracy on the noisy XOR test set is above $85\%$,
otherwise it is rejected.
Ensemble learning is terminated as soon as $1,000$ SGD solutions
with the required level of accuracy are obtained.

Figure
\ref{noisy_xor_parallel_coords_plot}
shows a parallel coordinates plot of
$100$
SGD solutions.
Each line
% represents an SGD solution,
connects the nine coordinates of a solution.
Overlaying lines of different SGD solutions
visualizes parameter symmetries.

Figure
\ref{noisy_xor_marginal_par_posteriors}
displays
% proxies for the marginal posterior of
histograms associated with
parameter $\theta_{3}$
of $\mbox{MLP}(2, 2, 1)$.
The green histogram represents
all $1,000$ SGD solutions for $\theta_{3}$
obtained from ensemble training
based on noisy XOR.
These $1,000$ modes cluster in two regions
approximately symmetric about zero.
The orange histogram belongs to
one of ten realized MH chains for $\theta_{3}$
based on noisy XOR.
This realized chain is entrapped in a
local mode in the vicinity of $5$,
where the orange histogram concentrates its mass.
The overlaid green and orange histograms
show that MH sampling explores a region
of the marginal posterior of $\theta_{3}$
also explored by ensemble training.

The blue histogram in figure
\ref{noisy_xor_marginal_par_posteriors}
comes from a chain realization for $\theta_{3}$
using MH sampling to apply $\mbox{MLP}(2, 2, 1)$
to the four exact XOR data points.
The pink line in figure
\ref{noisy_xor_marginal_par_posteriors}
shows the marginal prior
$\pi(\theta_{3})=\mathcal{N}(0,\sigma^2=10)$.
Four data points are not sufficient to learn from them,
given that $\mbox{MLP}(2, 2, 1)$ has nine parameters.
For this reason,
the blue histogram coincides with the pink line,
which means that the marginal posterior $p(\theta_{3})$
obtained from exact XOR via MH sampling
and the marginal prior $\pi (\theta_{3})$ coincide.

\subsection{Visual summaries for predictions}

Visual summaries for MLP predictions
and for MLP posterior predictive probabilities are
presented in this section.
MLP posterior predictive probabilities are visually shown
to quantify predictive uncertainty
in classification.

\subsubsection{Predictive accuracy.}

\begin{figure}[t]
		\centering
		\includegraphics[width=1\linewidth]{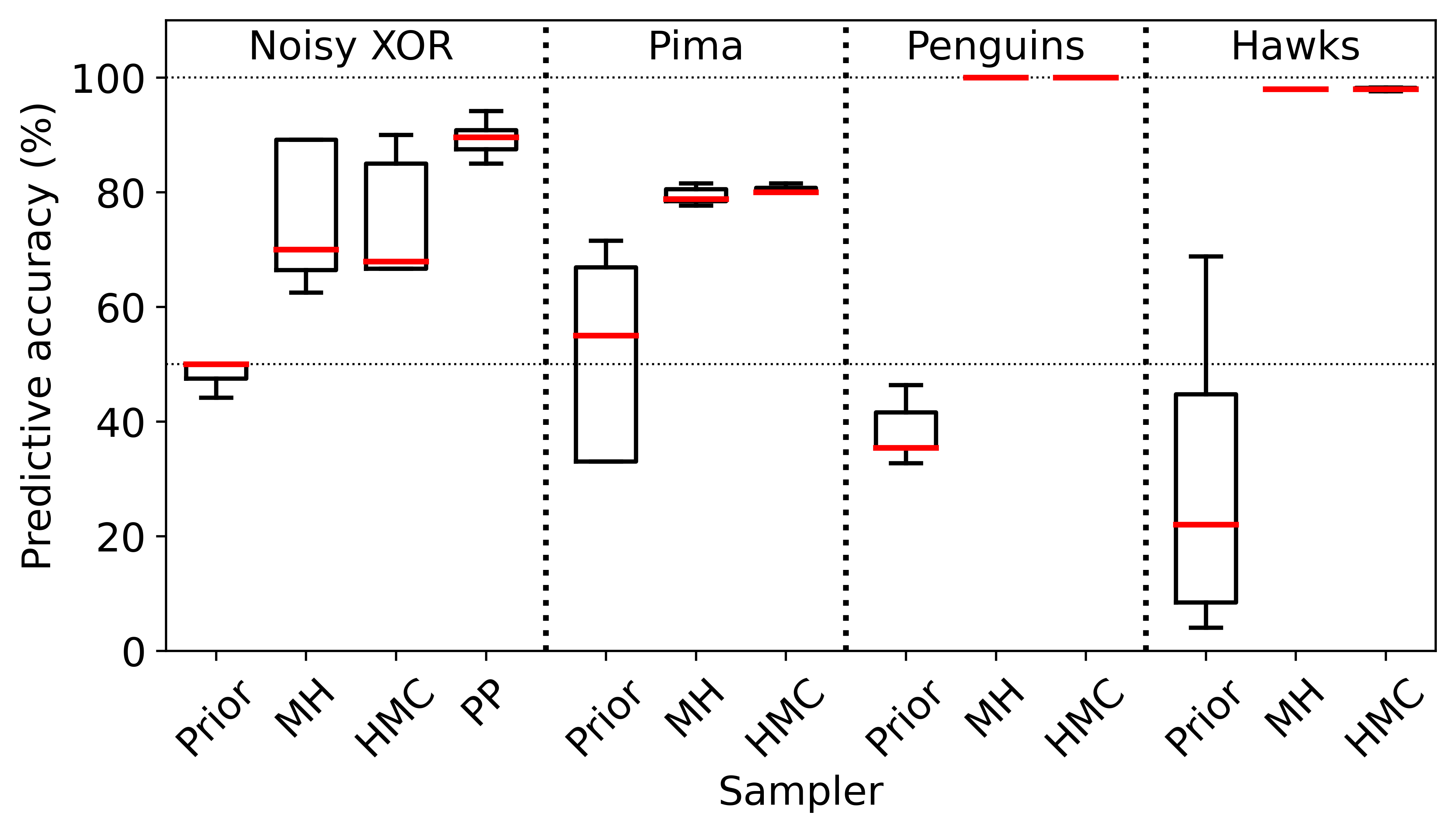}
	\caption{Boxplots of predictive accuracies
	for the examples introduced in table \ref{data_models_table}.
	Each boxplot summarizes $m=10$ predictive accuracies
    associated with the ten chains realized by an MCMC sampler
    for a test set.}
	\label{pred_boxplots}
\end{figure}

Figure \ref{pred_boxplots} shows boxplots of predictive accuracies,
hereinafter referred to as accuracies,
for the examples introduced in table \ref{data_models_table}.
Each boxplot summarizes $m=10$ accuracies
associated with the ten chains realized per sampler
for a test set.
Accuracy computation is based on Bayesian marginalization,
as outlined in subsections
\ref{subsec_bm} and \ref{subsec_experim_conf}.
Horizontal red lines represent accuracy medians.
Figure
\ref{pred_boxplots}
and
table
\ref{data_models_table}
provide complementary summaries,
as they present respective quartiles and means
of accuracies across chains per sampler.

Boxplot medians show high accuracy on the
penguin and hawk test sets.
Moreover, narrow boxplots
indicate accuracies with small variation on the
penguin and hawk test sets.
Thereby,
Bayesian marginalization based on
non-converged chain realizations
attains high accuracy with small variability
on the two multiclass classification examples.

Figure
\ref{pred_boxplots}
also displays boxplots of accuracies
based on expected posterior predictive distribution approximation
\eqref{pred_posterior_wrt_prior}
with respect to the prior.
For all four test sets
and regardless of Markov chain convergence,
Bayesian marginalization
outperforms agnostic prior-based baseline
\eqref{pred_posterior_wrt_prior}.

The PP boxplot has more elevated median and is narrower
than its MH and HMC counterparts for the noisy XOR test set.
This implies that PP sampling attains
higher accuracy with smaller variation
than MH and HMC sampling on the noisy XOR test set.

\subsubsection{Uncertainty quantification on a grid.}

\begin{figure}[t]
	\centering
	\includegraphics[width=1\linewidth]{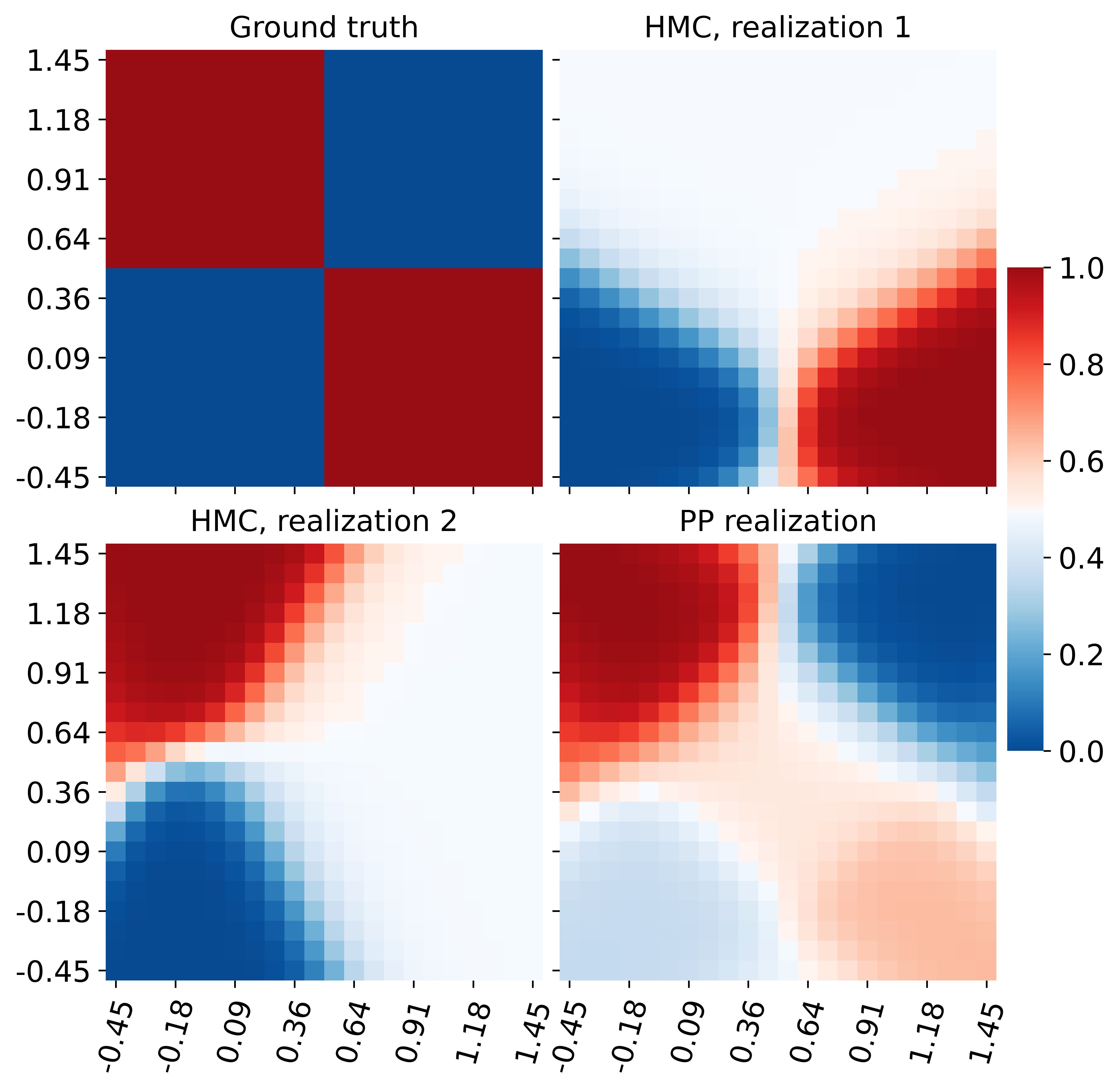}
	\caption{Heatmaps of ground truth
	and of posterior predictive probabilities
	$p(y=1 | (x_1, x_2), D_{1:500})=c$ on a grid
	of noisy XOR features $(x_1, x_2)$.
	The heatmap colour palette represents values of $c$.
	The ground truth heatmap visualizes true labels,
	while the other three heatmaps use approximate Bayesian
	marginalization based on HMC and PP chain realizations.}
	\label{pred_heatmaps}
\end{figure}

Figure \ref{pred_heatmaps} visualizes heatmaps
of the ground truth
and of posterior predictive distribution approximations
for noisy XOR.
More specifically, the posterior predictive probability
$p(y=1 | (x_1, x_2), D_{1:500})$
is approximated at the centre $(x_1, x_2)$ of each square cell
of a $22\times 22$ grid in $[-0.5, 1.5]\times [-0.5, 1.5]$.
$D_{1:500}$
refers to the noisy XOR training dataset of size
$s=500$
introduced in subsection
\ref{mlp_datasets}.
%Approximate BM on the basis of
%Monte Carlo integration X
%is conducted to evaluate
\eqref{pred_posterior_approx} is used for approximating
$p(y=1 | (x_1, x_2), D_{1:500})$.
Previously acquired
Markov chain realizations
(subsection \ref{num_summaries})
via MCMC sampling
of $\mbox{MLP}(2, 2, 1)$ parameters, using
the noisy XOR training dataset
$D_{1:500}$,
are passed to
\eqref{pred_posterior_approx}.

The approximation
$p(y=1 | (x_1, x_2), D_{1:500})=c$
at the center $(x_1, x_2)$ of a square cell
determines the colour of the cell in figure
\ref{pred_heatmaps}.
If $c$ is closer to $1$, $0$, or $0.5$,
the cell is plotted with a shade of
red, blue or white, respectively.
So, darker shades of red indicate
that $y=1$ with higher certainty,
darker shades of blue indicate
that $y=0$ with higher certainty,
and shades of white indicate
high uncertainty about
the binary label of noisy XOR.

Two posterior predictive distribution approximations
based on two HMC chain realizations
% are seen to
learn different regions of the
exact posterior predictive distribution.
Each of the two HMC chain realizations
uncover about half of the ground truth
of grid labels,
while it remains highly uncertain
for the other half of grid labels.
Moreover, both HMC chain realizations
exhibit higher uncertainty closer to the
decision boundaries
of ground truth.
These decision boundaries are
the vertical straight line $x_1=0.5$ and
horizontal straight line $x_2=0.5$.

A posterior predictive distribution approximation
based on a PP chain realization is displayed.
PP sampling uncovers larger regions of
the ground truth of grid labels than HMC sampling
in the considered grid
of noisy XOR features $(x_1, x_2)$.
Although HMC and PP samples do not converge to
the parameter posterior of $\mbox{MLP}(2, 2, 1)$,
approximate Bayesian marginalization
using these samples
predicts a subset of noisy XOR labels.

\subsubsection{Uncertainty quantification on a test set.}

\begin{figure}[t]
	\begin{subfigure}{.491\textwidth}
		\centering
		\includegraphics[width=0.5\linewidth]{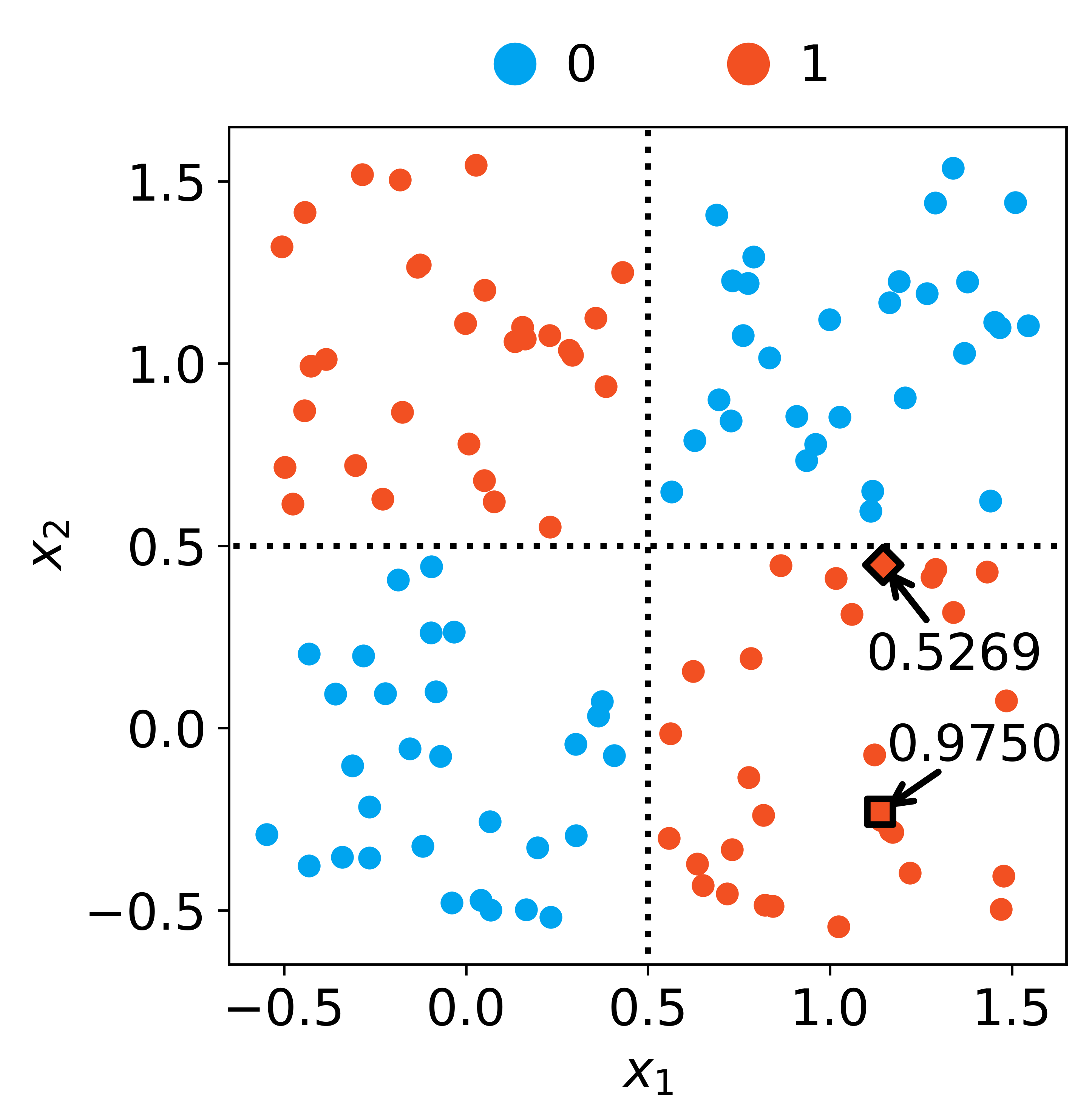}
		\caption{Scatterplot of noisy XOR features $(x_1, x_2)$.}
		\label{noisy_xor_scatter_uq}
	\end{subfigure}\\
	\begin{subfigure}{.491\textwidth}
	\centering
	\includegraphics[width=1\linewidth]{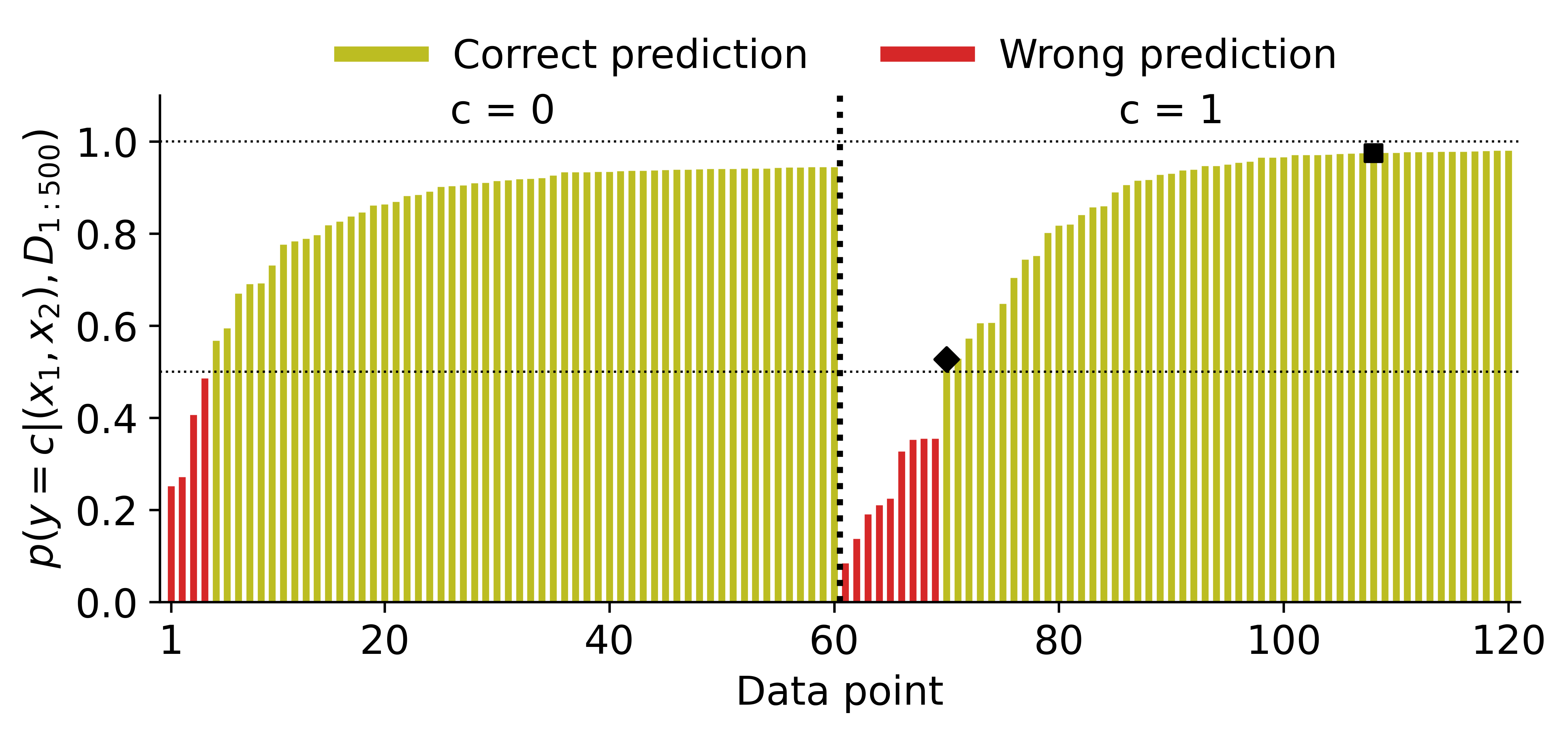}
	\caption{Posterior predictive probabilities for noisy XOR.}
	\label{noisy_xor_pred_posterior_uq_test}
	\end{subfigure}
	\caption{Quantification of uncertainty in predictions
	for the noisy XOR test set. Approximate Bayesian marginalization
	via MH sampling is used for computing posterior predictive probabilities.}
	\label{vis_summaries_uq_noisy_xor}
\end{figure}

\begin{figure}[t]
	\begin{subfigure}{.491\textwidth}
		\centering
		\includegraphics[width=1\linewidth]{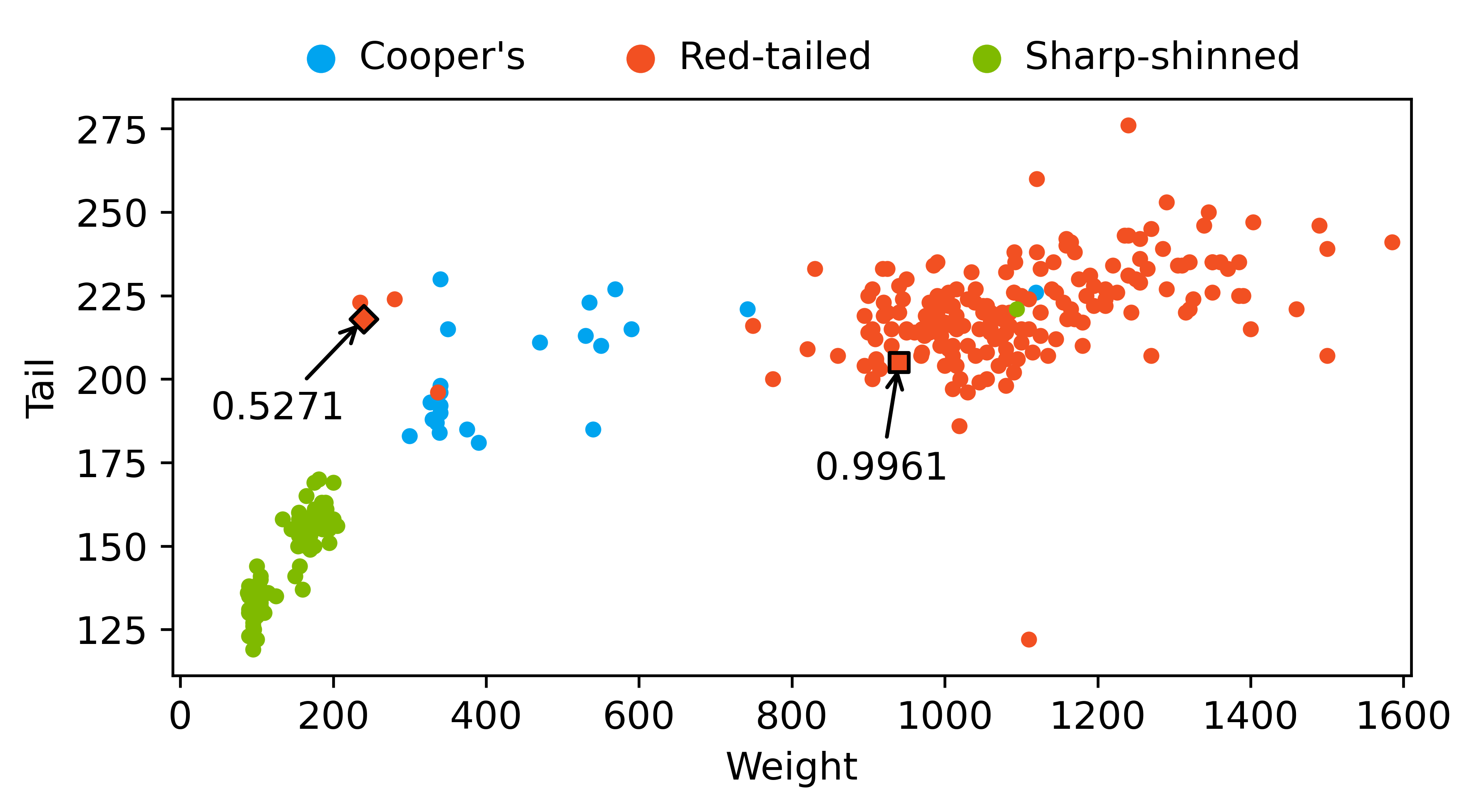}
		\caption{Scatterplot of hawks' weight against tail length.}
		\label{hawks_scatter_uq}
	\end{subfigure}\\
	\begin{subfigure}{.491\textwidth}
		\centering
		\includegraphics[width=1\linewidth]{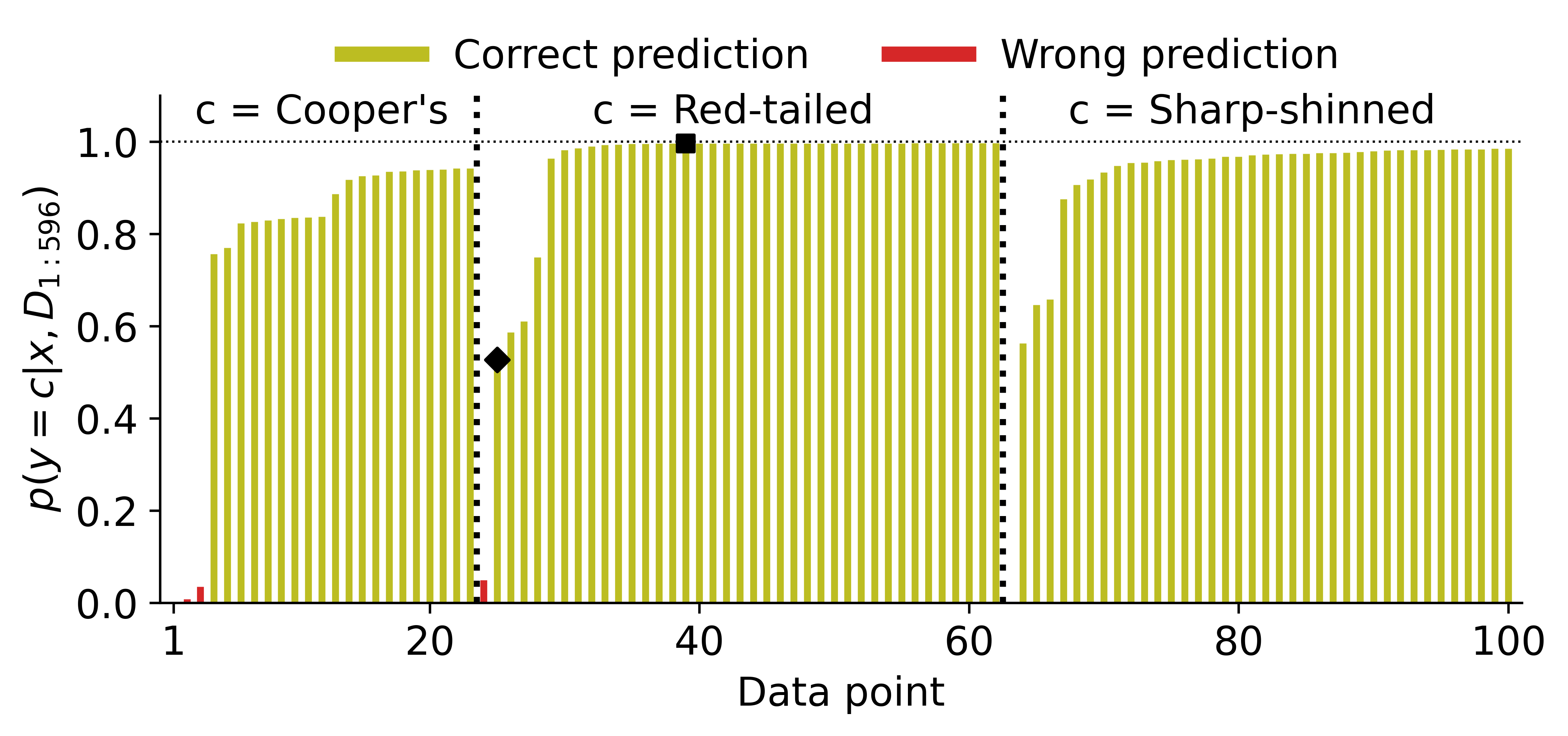}
		\caption{Posterior predictive probabilities for hawks.}
		\label{hawks_pred_posterior_uq_test}
	\end{subfigure}
	\caption{Quantification of uncertainty in predictions
	for the hawk test set. Bayesian marginalization
	via MH sampling is used for
	approximating posterior predictive probabilities.}
	\label{vis_summaries_uq_hawks}
\end{figure}

Figures
\ref{vis_summaries_uq_noisy_xor}
and
\ref{vis_summaries_uq_hawks}
show approximations of predictive prosterior probabilities
for a binary classification (noisy XOR) and
a multiclass classification (hawks) example.
Two posterior predictive probabilities
are interpreted contextually in each example
to quantify predictive uncertainty.

Figure
\ref{noisy_xor_scatter_uq}
visualizes the noisy XOR test set
of subsection
\ref{mlp_datasets}.
This is the same test set shown in figure
\ref{noisy_xor_scatterplots},
but with test points coloured according to their labels.
Figure
\ref{noisy_xor_pred_posterior_uq_test}
shows the posterior predictive probability
$p(y=c | (x_1, x_2), D_{1:500})$ of
true label $c\in\{0,1\}$
for each noisy XOR test point
$((x_1, x_2), y=c)$
given noisy XOR training set $D_{1:500}$
of subsection \ref{mlp_datasets}.
%thus, all predictive posterior probabilities
%of making correct predictions are displayed.
The posterior probabilities
$p(y=c | (x_1, x_2), D_{1:500})$
of predicting true class $c$
are ordered within class $c$.
Moreover,
each
$p(y=c | (x_1, x_2), D_{1:500})$
is coloured
as red or pale green depending on
whether the resulting prediction
is correct or not.
One of the ten MH chain realizations for
$\mbox{MLP}(2, 2, 1)$ parameter inference
from noisy XOR data
is used for approximating
$p(y=c | (x_1, x_2), D_{1:500})$
via
\eqref{pred_posterior_approx}
and for making predictions via
\eqref{bin_class_pred}.

Two points in the noisy XOR test set
are marked in figure
\ref{vis_summaries_uq_noisy_xor}
using a square and a rhombus.
These two points have the same true label $c=1$.
Given posterior predictive probabilities
$0.5269$
and
$0.9750$
for the rhombus and square-shaped test points,
the label $c=1$ is correctly predicted for both points.
However,
the rhombus-shaped point
is closer to
the decision boundary $x_2=0.5$
than the square-shaped point,
so classifying the former entails higher uncertainty.
As $0.5269 < 0.9750$,
Bayesian marginalization quantifies
the increased predictive uncertainty
associated with the rhombus-shaped point
despite using a non-converged MH chain realization.

Figure
\ref{hawks_scatter_uq}
shows a scatterplot of weight against tail length
for the hawk test set
of subsection
\ref{mlp_datasets}.
Blue, red and green test points belong to
Cooper's, red-tailed and sharp-shinned hawk classes.
Figure
\ref{hawks_pred_posterior_uq_test}
shows the posterior predictive probabilities
$p(y=c|x, D_{1:596})$
for a subset of $100$ hawk test points,
where
$c\in\{
\mbox{Cooper's},
\mbox{red-tailed},
\mbox{sharp-shinned}\}$
denotes the true label of test point
$(x, y=c)$
and
$D_{1:596}$
denotes the hawk training set
of subsection \ref{mlp_datasets}.
These posterior predictive probabilities
are shown ordered within each class,
and are coloured red or pale green
depending on whether they yield
correct or wrong predictions.
One of the ten MH chain realizations
for $\mbox{MLP}(6, 2, 2, 3)$ parameter inference
is used for approximating
$p(y=c | x, D_{1:596})$
via
\eqref{pred_posterior_approx}
and for making predictions via
\eqref{multi_class_pred}.

Two points in the hawk test set are marked in figure
\ref{vis_summaries_uq_hawks}
using a square and a rhombus.
Each of these two points represents weight and tail length
measurements from a red-tailed hawk.
The red-tailed hawk class is correctly predicted
for both points.
The squared-shaped observation belongs
to the main cluster of red-tailed hawks in figure
\ref{hawks_scatter_uq}
and it is predicted
with high posterior predictive probability
($0.9961$).
On the other hand,
the rhombus-shaped observation,
which falls in the cluster of Cooper's hawk,
is correctly predicted with a
lower posterior predictive probability ($0.5271$).
Bayesian marginalization
provides approximate posterior predictive probabilities
that signify the level of uncertainty in predictions
despite using a non-converged MH chain realization.

\subsection{Source code}

The source code for this paper is split into
three \texttt{Python} packages,
namely \texttt{eeyore}, \texttt{kanga} and \texttt{bnn\_mcmc\_examples}.
% A package called
\texttt{eeyore}
implements MCMC algorithms for Bayesian neural networks.
%A package called
\texttt{kanga}
implements MCMC diagnostics.
%A package called
\texttt{bnn\_mcmc\_examples}
includes the examples of this paper.

\texttt{eeyore} is available
via \texttt{pip},
via \texttt{conda}
and at
\url{https://github.com/papamarkou/eeyore}.
\texttt{eeyore} implements the MLP model,
as defined by \eqref{mlp_g}-\eqref{mlp_h},
using \texttt{PyTorch}.
An \texttt{MLP} class is set to be a subclass of
\texttt{torch.nn.Module},
with log-likelihood \eqref{bc_mlp_loglik} for binary classification
equal to the negative value of
\texttt{torch.nn.BCELoss} and
with log-likelihood \eqref{mc_mlp_loglik} for multiclass classification
equal to the negative value of \texttt{torch.nn.CrossEntropyLoss}.
Each MCMC algorithm takes an instance of \texttt{torch.nn.Module} as input,
with the logarithm of the target density being a
\texttt{log\_target} method of the instance.
Log-target density gradients for HMC are computed via
the automatic differentiation functionality of the
\texttt{torch.autograd} package of \texttt{PyTorch}.
The \texttt{MLP} class of \texttt{eeyore} provides a
\texttt{predictive\_posterior} method,
which implements the posterior predictive distribution approximation
\eqref{pred_posterior_approx}
given a realized Markov chain.

\texttt{kanga} is available
via \texttt{pip},
via \texttt{conda}
and at
\url{https://github.com/papamarkou/kanga}.
\texttt{kanga} is a collection of MCMC diagnostics
implemented using \texttt{numpy}.
MINSE, multivariate PSRF
% \eqref{eq_psrf} and
multivariate ESS
% \eqref{eq_ess}
are available in \texttt{kanga}.

\texttt{bnn\_mcmc\_examples} organizes the examples of this paper
in a package.
\texttt{bnn\_mcmc\_examples} relies on \texttt{eeyore} for
MCMC simulations and posterior predictive distribution approximations,
and on \texttt{kanga} for MCMC diagnostics.
For more details, see
\url{https://github.com/papamarkou/bnn_mcmc_examples}.

Optimization via SGD for the example involving
$\mbox{MLP}(2, 2, 1)$ and noisy XOR data
(figure \ref{vis_summaries_mcmc_vs_optim})
is run using \texttt{PyTorch}.
The loss function for optimization is computed via
\texttt{torch.nn.BCELoss}.
% and \texttt{torch.nn.CrossEntropyLoss}.
This loss function corresponds to the negative log-likelihood function
\eqref{bc_mlp_loglik}
% and \eqref{mc_mlp_loglik}
involved in MCMC,
thus linking the SGD and MH simulations
shown in figure \ref{noisy_xor_marginal_par_posteriors}.
SGD is coded manually
instead of calling an optimization algorithm
of the \texttt{torch.optim} package of \texttt{PyTorch}.
Gradients for optimization are computed calling the \texttt{backward} method.
The SGD code related to the example of
figure \ref{vis_summaries_mcmc_vs_optim}
is available at
\url{https://github.com/papamarkou/bnn_mcmc_examples}.

\subsection{Hardware}
\label{hardware}

Pilot MCMC runs
indicated an increase in speed by using CPUs instead of GPUs;
accordingly, computations were performed on CPUs for this paper.
The GPU slowdown is explained
by the overhead of copying \texttt{PyTorch} tensors between GPUs and CPUs
for small neural networks, such as the ones used in section \ref{examples}.

The computations for section \ref{examples}
were run on Google Cloud Platform (GCP).
Eleven virtual machine (VM) instances
with virtual CPUs % (vCPUs)
were created on GCP
to spread the workload.
% to allocate workload of independent experiments to different CPUs.

Setting aside heterogeneities in hardware configuration
between GCP VM instances
and in order to provide an indication of
% relative
computational cost,
% across different MCMC samplers,
MCMC simulation runtimes are provided for the example
of applying an $\mbox{MLP}(6, 2, 2, 3)$ to the hawk training dataset.
The mean runtimes across the ten realized chains
per MH and HMC
are
$0:42:54$ and
$1:10:48$,
respectively (runtimes are formatted as `hours : minutes : seconds').

\section{Predictive inference scope}
\label{scope}

Bayesian marginalization can attain high predictive accuracy
and can quantify predictive uncertainty
using non-converged MCMC samples of neural network parameters.
Thus,
MCMC sampling elicits some information
about the parameter posterior of a neural network
and conveys such information to the posterior predictive distribution.
It is possible that MCMC sampling
learns about the statistical dependence
among neural network parameters.
Along these lines,
groups of weights or biases can be formed,
with strong within-group and weak between-group dependence,
to investigate scalable block Gibbs sampling methods
for neural networks.

Another possibility of MCMC developments for neural networks
entails shifting attention from the parameter space to the output space,
since the latter is related to predictive inference directly.
Approximate MCMC methods
that measure the discrepancy or Wasserstein distance
between neural network predictions and output data \citep{rudolf2018}
can be investigated.

Bayesian marginalization provides scope
to develop predictive inference for neural networks.
For instance,
Bayesian marginalization can be examined in the context of
approximate MCMC sampling from a neural network parameter posterior,
regardless of convergence to the parameter posterior and
in analogy to the workings of this paper.
Moreover, the idea of \citet{wilson2020} to interpret
ensemble training of neural networks from
a viewpoint of Bayesian marginalization
can be studied using the notion of quantization
of probability distributions.

\section*{Appendix A: power posteriors}
\sname{Appendix A}
\label{appendix_categorical}

% \subsection{Categorical distribution for power posterior state swaps}

This appendix provides the
probability mass function $p_i(j)$
for proposing a chain $j$
for a possible swap of states between chains $i$ and $j$
in PP sampling.
Assuming $m+1$ power posteriors, a neighbouring chain $j$ of $i$ is chosen
randomly from the categorical probability mass function
$p_i =
\mathcal{C}(\alpha_i(0), \alpha_i(1),\dots,
\alpha_i(i-1),\alpha_i(i+1),\dots,\alpha_i(m))$
with event probabilities
\begin{equation*}
\alpha_i(j) = \frac{\exp{(-\beta |j-i|)}}{\gamma_i},
\end{equation*}
where $i\in\{0,1,\dots,m\}$,
$j\in\{0,1,\dots,m\}\setminus\{i\}$,
$\beta$ is a hyperparameter and
$\gamma_i$ is a normalizing constant.
The hyperparameter $\beta$ is typically set to $\beta=0.5$,
a value which makes a jump to $j=i\pm 1$ roughly three times more likely than a
jump to $j=i\pm 3$ \citep{friel2008}.

The normalizing constant $\gamma_i$ is given by
\begin{equation*}
\gamma_i =
\frac{\exp{(-\beta)}(2-\exp{(-\beta i)}-\exp{(-\beta
(m-i))})}{1-\exp{(-\beta)}}.
\end{equation*}
Starting from the fact that the event probabilities $\alpha_i(j)$ add up to one,
$\gamma_i$ is derived as follows:
\begin{equation*}
\begin{split}
1 &=
\sum_{\substack{j=0\\ j\ne i}}^{m}\alpha_i (j)\Rightarrow\\
\gamma_i &=
\sum_{j=0}^{i-1}\exp{(-\beta(i-j))}+\sum_{j=i+1}^{m}\exp{(-\beta (j-i))}\\
&=
\sum_{j=1}^{i}\exp{(-\beta j)}+\sum_{j=1}^{m-i}\exp{(-\beta j)}\\
&=
\exp{(-\beta)}\left(\frac{1-\exp{(-\beta i)}}{1-\exp{(-\beta)}}\right)\\
&
+\exp{(-\beta)}\left(\frac{1-\exp{(-\beta (m-i))}}{1-\exp{(-\beta)}}\right)\\
&=
\frac{\exp{(-\beta)}(2-\exp{(-\beta i)}-\exp{(-\beta (m-i))})}{1-\exp{(-\beta)}}.
\end{split}
\end{equation*}

\section*{Appendix B: Predictive distribution}
\sname{Appendix B}
\label{appendix_predictive}

This appendix derives the posterior predictive distribution \eqref{pred_posterior}.
Applying the law of total probability and
the definition of conditional probability yields
%\begin{align}
\begin{equation*}
\begin{split}
p(y | x, D_{1:s})
& =
\int p(y, \theta | x, D_{1:s}) d\theta\\
& =
\int p(y | x, D_{1:s}, \theta)
p(\theta | x, D_{1:s})
d\theta .
\end{split}
\end{equation*}
%\end{align}
%$y$ and $D_{1:s}$ are conditionally independent
%given $(x,\theta)$, since
$p(y | x, D_{1:s}, \theta)$ is equal to
the likelihood $p(y | x, \theta)$:
\begin{equation*}
\begin{split}
p(y | x, D_{1:s}, \theta)
& =
\frac{p(y, D_{1:s} | x, \theta)}{p(D_{1:s} | x, \theta)}\\
& =
\frac{p(y | x, \theta) p(D_{1:s} | x, \theta)}
{p(D_{1:s} | x, \theta)}\\
& =
p(y | x, \theta) .
\end{split}
\end{equation*}
%$\theta$ and $x$ are conditionally independent
%given $D_{1:s}$, since
Furthermore, $p(\theta | x, D_{1:s})$ is equal to
the parameter posterior $p(\theta | D_{1:s})$:
\begin{equation*}
\begin{split}
p(\theta | x, D_{1:s})
& =
\frac{p(\theta, x, D_{1:s})}{p(x, D_{1:s})}\\
& =
\frac{p(\theta, x | D_{1:s}) p(D_{1:s})}
{p(x) p(D_{1:s})}\\
& =
\frac{p(\theta | D_{1:s}) p(x | D_{1:s})}
{p(x)}\\
& =
\frac{p(\theta | D_{1:s}) p(x, D_{1:s})}
{p(x)p(D_{1:s})}\\
& =
\frac{p(\theta | D_{1:s}) p(x) p(D_{1:s})}
{p(x)p(D_{1:s})}\\
& =
p(\theta | D_{1:s}).
\end{split}
\end{equation*}
% thus completing the derivation of \eqref{pred_posterior}.

\section*{Acknowledgements}

Research sponsored by the Laboratory Directed Research and Development Program
of Oak Ridge National Laboratory,
managed by UT-Battelle, LLC,
for the US Department of Energy under contract DE-AC05-00OR22725.

The first author would like to thank Google
for the provision of free credit on Google Cloud Platform.

%This research used resources of
%the Compute and Data Environment for Science (CADES) at the Oak Ridge National Laboratory,
%which is supported by the Office of Science of the U.S. Department of Energy
% under Contract No. DE-AC05-00OR22725.

% \bibliographystyle{imsart-nameyear}
% \bibliography{references}

\begin{thebibliography}{101}
% BibTex style file: imsart-nameyear.bst, 2017-11-03
% Default style options (sort=1,type=nameyear).
% Used options (sort=1,type=nameyear).

\bibitem[\protect\citeauthoryear{Andrieu, de~Freitas and
  Doucet}{1999}]{andrieu1999}
\begin{bmisc}[author]
\bauthor{\bsnm{Andrieu},~\bfnm{C.}\binits{C.}}, \bauthor{\bparticle{de}
  \bsnm{Freitas},~\bfnm{J.~F.~G.}\binits{J.~F.~G.}} \AND
  \bauthor{\bsnm{Doucet},~\bfnm{A.}\binits{A.}}
(\byear{1999}).
\btitle{Sequential {B}ayesian estimation and model selection applied to neural
  networks}.
\end{bmisc}
\endbibitem

\bibitem[\protect\citeauthoryear{Andrieu, de~Freitas and
  Doucet}{2000}]{andrieu2000}
\begin{binproceedings}[author]
\bauthor{\bsnm{Andrieu},~\bfnm{Christophe}\binits{C.}}, \bauthor{\bparticle{de}
  \bsnm{Freitas},~\bfnm{Nando}\binits{N.}} \AND
  \bauthor{\bsnm{Doucet},~\bfnm{Arnaud}\binits{A.}}
(\byear{2000}).
\btitle{Reversible jump {MCMC} simulated annealing for neural networks}.
In \bbooktitle{Proceedings of the Sixteenth Conference on Uncertainty in
  Artificial Intelligence}
\bpages{11--18}.
\end{binproceedings}
\endbibitem

\bibitem[\protect\citeauthoryear{Ashukha et~al.}{2020}]{ashukha2020}
\begin{binproceedings}[author]
\bauthor{\bsnm{Ashukha},~\bfnm{Arsenii}\binits{A.}},
  \bauthor{\bsnm{Lyzhov},~\bfnm{Alexander}\binits{A.}},
  \bauthor{\bsnm{Molchanov},~\bfnm{Dmitry}\binits{D.}} \AND
  \bauthor{\bsnm{Vetrov},~\bfnm{Dmitry}\binits{D.}}
(\byear{2020}).
\btitle{Pitfalls of in-domain uncertainty estimation and ensembling in deep
  learning}.
In \bbooktitle{International Conference on Learning Representations}.
\end{binproceedings}
\endbibitem

\bibitem[\protect\citeauthoryear{Badrinarayanan, Mishra and
  Cipolla}{2015}]{badrinarayanan2015}
\begin{barticle}[author]
\bauthor{\bsnm{Badrinarayanan},~\bfnm{Vijay}\binits{V.}},
  \bauthor{\bsnm{Mishra},~\bfnm{Bamdev}\binits{B.}} \AND
  \bauthor{\bsnm{Cipolla},~\bfnm{Roberto}\binits{R.}}
(\byear{2015}).
\btitle{Symmetry-invariant optimization in deep networks}.
\bjournal{arXiv}.
\end{barticle}
\endbibitem

\bibitem[\protect\citeauthoryear{Bennett, Racine-Poon and
  Wakefield}{}]{bennett1996}
\begin{bincollection}[author]
\bauthor{\bsnm{Bennett},~\bfnm{James~E}\binits{J.~E.}},
  \bauthor{\bsnm{Racine-Poon},~\bfnm{Amy}\binits{A.}} \AND
  \bauthor{\bsnm{Wakefield},~\bfnm{Jon~C}\binits{J.~C.}}
\btitle{{MCMC} for nonlinear hierarchical models}.
\bpages{339--358}.
\end{bincollection}
\endbibitem

\bibitem[\protect\citeauthoryear{Bernardo}{1979}]{bernardo1979}
\begin{barticle}[author]
\bauthor{\bsnm{Bernardo},~\bfnm{Jose~M.}\binits{J.~M.}}
(\byear{1979}).
\btitle{Reference posterior distributions for {B}ayesian inference}.
\bjournal{Journal of the Royal Statistical Society. Series B (Methodological)}
\bvolume{41}
\bpages{113--147}.
\end{barticle}
\endbibitem

\bibitem[\protect\citeauthoryear{Blei, Kucukelbir and
  McAuliffe}{2017}]{blei2017}
\begin{barticle}[author]
\bauthor{\bsnm{Blei},~\bfnm{David~M.}\binits{D.~M.}},
  \bauthor{\bsnm{Kucukelbir},~\bfnm{Alp}\binits{A.}} \AND
  \bauthor{\bsnm{McAuliffe},~\bfnm{Jon~D.}\binits{J.~D.}}
(\byear{2017}).
\btitle{Variational inference: a review for statisticians}.
\bjournal{Journal of the American Statistical Association}
\bvolume{112}
\bpages{859-877}.
\end{barticle}
\endbibitem

\bibitem[\protect\citeauthoryear{Blier and Ollivier}{2018}]{blier2018}
\begin{binproceedings}[author]
\bauthor{\bsnm{Blier},~\bfnm{L\'{e}onard}\binits{L.}} \AND
  \bauthor{\bsnm{Ollivier},~\bfnm{Yann}\binits{Y.}}
(\byear{2018}).
\btitle{The description length of deep learning models}.
In \bbooktitle{Advances in Neural Information Processing Systems}
\bvolume{31}.
\end{binproceedings}
\endbibitem

\bibitem[\protect\citeauthoryear{Brea et~al.}{2019}]{brea2019}
\begin{barticle}[author]
\bauthor{\bsnm{Brea},~\bfnm{Johanni}\binits{J.}},
  \bauthor{\bsnm{Simsek},~\bfnm{Berfin}\binits{B.}},
  \bauthor{\bsnm{Illing},~\bfnm{Bernd}\binits{B.}} \AND
  \bauthor{\bsnm{Gerstner},~\bfnm{Wulfram}\binits{W.}}
(\byear{2019}).
\btitle{Weight-space symmetry in deep networks gives rise to permutation
  saddles, connected by equal-loss valleys across the loss landscape}.
\bjournal{arXiv}.
\end{barticle}
\endbibitem

\bibitem[\protect\citeauthoryear{Brooks and Gelman}{1998}]{brooks1998}
\begin{barticle}[author]
\bauthor{\bsnm{Brooks},~\bfnm{Stephen~P.}\binits{S.~P.}} \AND
  \bauthor{\bsnm{Gelman},~\bfnm{Andrew}\binits{A.}}
(\byear{1998}).
\btitle{General methods for monitoring convergence of iterative simulations}.
\bjournal{Journal of Computational and Graphical Statistics}
\bvolume{7}
\bpages{434--455}.
\end{barticle}
\endbibitem

\bibitem[\protect\citeauthoryear{Cannon et~al.}{2019}]{cannon2019}
\begin{bmanual}[author]
\bauthor{\bsnm{Cannon},~\bfnm{Ann}\binits{A.}},
  \bauthor{\bsnm{Cobb},~\bfnm{George}\binits{G.}},
  \bauthor{\bsnm{Hartlaub},~\bfnm{Bradley}\binits{B.}},
  \bauthor{\bsnm{Legler},~\bfnm{Julie}\binits{J.}},
  \bauthor{\bsnm{Lock},~\bfnm{Robin}\binits{R.}},
  \bauthor{\bsnm{Moore},~\bfnm{Thomas}\binits{T.}},
  \bauthor{\bsnm{Rossman},~\bfnm{Allan}\binits{A.}} \AND
  \bauthor{\bsnm{Witmer},~\bfnm{Jeffrey}\binits{J.}}
(\byear{2019}).
\btitle{Stat2{D}ata: datasets for {S}tat2}
\bnote{R package version 2.0.0}.
\end{bmanual}
\endbibitem

\bibitem[\protect\citeauthoryear{Chen, Fox and Guestrin}{2014}]{chen2014}
\begin{binproceedings}[author]
\bauthor{\bsnm{Chen},~\bfnm{Tianqi}\binits{T.}},
  \bauthor{\bsnm{Fox},~\bfnm{Emily}\binits{E.}} \AND
  \bauthor{\bsnm{Guestrin},~\bfnm{Carlos}\binits{C.}}
(\byear{2014}).
\btitle{Stochastic gradient {H}amiltonian {M}onte {C}arlo}.
In \bbooktitle{Proceedings of the 31st International Conference on Machine
  Learning}
\bvolume{32}
\bpages{1683--1691}.
\end{binproceedings}
\endbibitem

\bibitem[\protect\citeauthoryear{Chen, Lu and Hecht-Nielsen}{1993}]{chen1993}
\begin{barticle}[author]
\bauthor{\bsnm{Chen},~\bfnm{A.~M.}\binits{A.~M.}},
  \bauthor{\bsnm{Lu},~\bfnm{H.}\binits{H.}} \AND
  \bauthor{\bsnm{Hecht-Nielsen},~\bfnm{R.}\binits{R.}}
(\byear{1993}).
\btitle{On the geometry of feedforward neural network error surfaces}.
\bjournal{Neural Computation}
\bvolume{5}
\bpages{910--927}.
\end{barticle}
\endbibitem

\bibitem[\protect\citeauthoryear{Chen et~al.}{2019}]{chen2019}
\begin{binproceedings}[author]
\bauthor{\bsnm{Chen},~\bfnm{Wilson~Ye}\binits{W.~Y.}},
  \bauthor{\bsnm{Barp},~\bfnm{Alessandro}\binits{A.}},
  \bauthor{\bsnm{Briol},~\bfnm{Francois-Xavier}\binits{F.-X.}},
  \bauthor{\bsnm{Gorham},~\bfnm{Jackson}\binits{J.}},
  \bauthor{\bsnm{Girolami},~\bfnm{Mark}\binits{M.}},
  \bauthor{\bsnm{Mackey},~\bfnm{Lester}\binits{L.}} \AND
  \bauthor{\bsnm{Oates},~\bfnm{Chris}\binits{C.}}
(\byear{2019}).
\btitle{Stein point {M}arkov chain {M}onte {C}arlo}.
In \bbooktitle{Proceedings of the 36th International Conference on Machine
  Learning}
\bvolume{97}
\bpages{1011--1021}.
\end{binproceedings}
\endbibitem

\bibitem[\protect\citeauthoryear{Chollet}{2017}]{chollet2017}
\begin{binproceedings}[author]
\bauthor{\bsnm{Chollet},~\bfnm{Fran{\c{c}}ois}\binits{F.}}
(\byear{2017}).
\btitle{{X}ception: deep learning with depthwise separable convolutions}.
In \bbooktitle{Proceedings of the IEEE conference on computer vision and
  pattern recognition}
\bpages{1251--1258}.
\end{binproceedings}
\endbibitem

\bibitem[\protect\citeauthoryear{Chwialkowski, Strathmann and
  Gretton}{2016}]{chwialkowski2016}
\begin{binproceedings}[author]
\bauthor{\bsnm{Chwialkowski},~\bfnm{Kacper}\binits{K.}},
  \bauthor{\bsnm{Strathmann},~\bfnm{Heiko}\binits{H.}} \AND
  \bauthor{\bsnm{Gretton},~\bfnm{Arthur}\binits{A.}}
(\byear{2016}).
\btitle{A kernel test of goodness of fit}.
In \bbooktitle{Proceedings of The 33rd International Conference on Machine
  Learning}
\bvolume{48}
\bpages{2606--2615}.
\end{binproceedings}
\endbibitem

\bibitem[\protect\citeauthoryear{Cowles and Carlin}{1996}]{cowles1996}
\begin{barticle}[author]
\bauthor{\bsnm{Cowles},~\bfnm{Mary~Kathryn}\binits{M.~K.}} \AND
  \bauthor{\bsnm{Carlin},~\bfnm{Bradley~P.}\binits{B.~P.}}
(\byear{1996}).
\btitle{Markov chain {M}onte {C}arlo convergence diagnostics: a comparative
  review}.
\bjournal{Journal of the American Statistical Association}
\bvolume{91}
\bpages{883--904}.
\end{barticle}
\endbibitem

\bibitem[\protect\citeauthoryear{Cybenko}{1989}]{cybenko1989}
\begin{barticle}[author]
\bauthor{\bsnm{Cybenko},~\bfnm{George}\binits{G.}}
(\byear{1989}).
\btitle{Approximation by superpositions of a sigmoidal function}.
\bjournal{Mathematics of control, signals and systems}
\bvolume{2}
\bpages{303--314}.
\end{barticle}
\endbibitem

\bibitem[\protect\citeauthoryear{Dai and Jones}{2017}]{dai2017}
\begin{barticle}[author]
\bauthor{\bsnm{Dai},~\bfnm{Ning}\binits{N.}} \AND
  \bauthor{\bsnm{Jones},~\bfnm{Galin~L.}\binits{G.~L.}}
(\byear{2017}).
\btitle{Multivariate initial sequence estimators in {M}arkov chain {M}onte
  {C}arlo}.
\bjournal{Journal of Multivariate Analysis}
\bvolume{159}
\bpages{184--199}.
\end{barticle}
\endbibitem

\bibitem[\protect\citeauthoryear{Daniels and Kass}{1998}]{daniels1998}
\begin{barticle}[author]
\bauthor{\bsnm{Daniels},~\bfnm{Michael~J.}\binits{M.~J.}} \AND
  \bauthor{\bsnm{Kass},~\bfnm{Robert~E.}\binits{R.~E.}}
(\byear{1998}).
\btitle{A note on first-stage approximation in two-stage hierarchical models}.
\bjournal{Sankhy\={a}: The Indian Journal of Statistics, Series B (1960-2002)}
\bvolume{60}
\bpages{19--30}.
\end{barticle}
\endbibitem

\bibitem[\protect\citeauthoryear{de~Freitas}{1999}]{freitas1999}
\begin{bphdthesis}[author]
\bauthor{\bparticle{de} \bsnm{Freitas},~\bfnm{Nando}\binits{N.}}
(\byear{1999}).
\btitle{Bayesian methods for neural networks},
\btype{PhD thesis},
\bpublisher{University of Cambridge}.
\end{bphdthesis}
\endbibitem

\bibitem[\protect\citeauthoryear{de~Freitas et~al.}{2001}]{freitas2001}
\begin{binbook}[author]
\bauthor{\bparticle{de} \bsnm{Freitas},~\bfnm{N.}\binits{N.}},
  \bauthor{\bsnm{Andrieu},~\bfnm{C.}\binits{C.}},
  \bauthor{\bsnm{H{\o}jen-S{\o}rensen},~\bfnm{P.}\binits{P.}},
  \bauthor{\bsnm{Niranjan},~\bfnm{M.}\binits{M.}} \AND
  \bauthor{\bsnm{Gee},~\bfnm{A.}\binits{A.}}
(\byear{2001}).
\btitle{Sequential {M}onte {C}arlo methods for neural networks}
In \bbooktitle{Sequential Monte Carlo Methods in Practice}
\bpages{359--379}.
\end{binbook}
\endbibitem

\bibitem[\protect\citeauthoryear{De~Sa, Chen and Wong}{2018}]{desa2018}
\begin{binproceedings}[author]
\bauthor{\bsnm{De~Sa},~\bfnm{Chris}\binits{C.}},
  \bauthor{\bsnm{Chen},~\bfnm{Vincent}\binits{V.}} \AND
  \bauthor{\bsnm{Wong},~\bfnm{Wing}\binits{W.}}
(\byear{2018}).
\btitle{Minibatch {G}ibbs sampling on large graphical models}.
In \bbooktitle{Proceedings of the 35th International Conference on Machine
  Learning}
\bvolume{80}
\bpages{1165--1173}.
\end{binproceedings}
\endbibitem

\bibitem[\protect\citeauthoryear{Dupuy and Bach}{2017}]{dupuy2017}
\begin{barticle}[author]
\bauthor{\bsnm{Dupuy},~\bfnm{Christophe}\binits{C.}} \AND
  \bauthor{\bsnm{Bach},~\bfnm{Francis}\binits{F.}}
(\byear{2017}).
\btitle{Online but accurate inference for latent variable models with local
  {G}ibbs sampling}.
\bjournal{Journal of Machine Learning Research}
\bvolume{18}
\bpages{1-45}.
\end{barticle}
\endbibitem

\bibitem[\protect\citeauthoryear{Ensign et~al.}{2017}]{ensign2017}
\begin{binproceedings}[author]
\bauthor{\bsnm{Ensign},~\bfnm{Danielle}\binits{D.}},
  \bauthor{\bsnm{Neville},~\bfnm{Scott}\binits{S.}},
  \bauthor{\bsnm{Paul},~\bfnm{Arnab}\binits{A.}} \AND
  \bauthor{\bsnm{Venkatasubramanian},~\bfnm{Suresh}\binits{S.}}
(\byear{2017}).
\btitle{The complexity of explaining neural networks through (group)
  invariants}.
In \bbooktitle{Proceedings of the 28th International Conference on Algorithmic
  Learning Theory}
\bvolume{76}
\bpages{341--359}.
\end{binproceedings}
\endbibitem

\bibitem[\protect\citeauthoryear{Esmaeili et~al.}{2019}]{esmaeili2019}
\begin{binproceedings}[author]
\bauthor{\bsnm{Esmaeili},~\bfnm{Babak}\binits{B.}},
  \bauthor{\bsnm{Wu},~\bfnm{Hao}\binits{H.}},
  \bauthor{\bsnm{Jain},~\bfnm{Sarthak}\binits{S.}},
  \bauthor{\bsnm{Bozkurt},~\bfnm{Alican}\binits{A.}},
  \bauthor{\bsnm{Siddharth},~\bfnm{N}\binits{N.}},
  \bauthor{\bsnm{Paige},~\bfnm{Brooks}\binits{B.}},
  \bauthor{\bsnm{Brooks},~\bfnm{Dana~H.}\binits{D.~H.}},
  \bauthor{\bsnm{Dy},~\bfnm{Jennifer}\binits{J.}} \AND
  \bauthor{\bparticle{van~de} \bsnm{Meent},~\bfnm{Jan-Willem}\binits{J.-W.}}
(\byear{2019}).
\btitle{Structured disentangled representations}.
In \bbooktitle{Proceedings of the 22nd International Conference on Artificial
  Intelligence and Statistics}
\bvolume{89}
\bpages{2525--2534}.
\end{binproceedings}
\endbibitem

\bibitem[\protect\citeauthoryear{Freeman, Roese-Koerner and
  Kummert}{2018}]{freeman2018}
\begin{binproceedings}[author]
\bauthor{\bsnm{Freeman},~\bfnm{Ido}\binits{I.}},
  \bauthor{\bsnm{Roese-Koerner},~\bfnm{Lutz}\binits{L.}} \AND
  \bauthor{\bsnm{Kummert},~\bfnm{Anton}\binits{A.}}
(\byear{2018}).
\btitle{Effnet: An efficient structure for convolutional neural networks}.
In \bbooktitle{25th IEEE International Conference on Image Processing}
\bpages{6--10}.
\end{binproceedings}
\endbibitem

\bibitem[\protect\citeauthoryear{Friel and Pettitt}{2008}]{friel2008}
\begin{barticle}[author]
\bauthor{\bsnm{Friel},~\bfnm{N.}\binits{N.}} \AND
  \bauthor{\bsnm{Pettitt},~\bfnm{A.~N.}\binits{A.~N.}}
(\byear{2008}).
\btitle{Marginal likelihood estimation via power posteriors}.
\bjournal{Journal of the Royal Statistical Society: Series B (Statistical
  Methodology)}
\bvolume{70}
\bpages{589--607}.
\end{barticle}
\endbibitem

\bibitem[\protect\citeauthoryear{Gelman and Rubin}{1992}]{gelman1992}
\begin{barticle}[author]
\bauthor{\bsnm{Gelman},~\bfnm{Andrew}\binits{A.}} \AND
  \bauthor{\bsnm{Rubin},~\bfnm{Donald~B.}\binits{D.~B.}}
(\byear{1992}).
\btitle{Inference from iterative simulation using multiple sequences}.
\bjournal{Statistical Science}
\bvolume{7}
\bpages{457--472}.
\end{barticle}
\endbibitem

\bibitem[\protect\citeauthoryear{Gelman et~al.}{2004}]{gelman2004}
\begin{bbook}[author]
\bauthor{\bsnm{Gelman},~\bfnm{Andrew}\binits{A.}},
  \bauthor{\bsnm{Carlin},~\bfnm{John~B}\binits{J.~B.}},
  \bauthor{\bsnm{Stern},~\bfnm{Hal~S}\binits{H.~S.}} \AND
  \bauthor{\bsnm{Rubin},~\bfnm{Donald~B}\binits{D.~B.}}
(\byear{2004}).
\btitle{Bayesian data analysis},
\bedition{2nd} ed.
\bpublisher{Chapman and Hall/CRC}.
\end{bbook}
\endbibitem

\bibitem[\protect\citeauthoryear{Gilks and Roberts}{}]{gilks1996b}
\begin{bincollection}[author]
\bauthor{\bsnm{Gilks},~\bfnm{Walter~R}\binits{W.~R.}} \AND
  \bauthor{\bsnm{Roberts},~\bfnm{Gareth~O}\binits{G.~O.}}
\btitle{Strategies for improving {MCMC}}.
\bpages{89--114}.
\end{bincollection}
\endbibitem

\bibitem[\protect\citeauthoryear{Giordano, Broderick and
  Jordan}{2015}]{giordano2015}
\begin{bincollection}[author]
\bauthor{\bsnm{Giordano},~\bfnm{Ryan~J}\binits{R.~J.}},
  \bauthor{\bsnm{Broderick},~\bfnm{Tamara}\binits{T.}} \AND
  \bauthor{\bsnm{Jordan},~\bfnm{Michael~I}\binits{M.~I.}}
(\byear{2015}).
\btitle{Linear response methods for accurate covariance estimates from mean
  field variational {B}ayes}.
In \bbooktitle{Advances in Neural Information Processing Systems 28}
\bpages{1441--1449}.
\end{bincollection}
\endbibitem

\bibitem[\protect\citeauthoryear{Gong and Flegal}{2016}]{gong2016}
\begin{barticle}[author]
\bauthor{\bsnm{Gong},~\bfnm{Lei}\binits{L.}} \AND
  \bauthor{\bsnm{Flegal},~\bfnm{James~M.}\binits{J.~M.}}
(\byear{2016}).
\btitle{A practical sequential stopping rule for high-dimensional {M}arkov
  chain {M}onte {C}arlo}.
\bjournal{Journal of Computational and Graphical Statistics}
\bvolume{25}
\bpages{684--700}.
\end{barticle}
\endbibitem

\bibitem[\protect\citeauthoryear{Gong, Li and
  Hern\'{a}ndez-Lobato}{2019}]{gong2019}
\begin{binproceedings}[author]
\bauthor{\bsnm{Gong},~\bfnm{Wenbo}\binits{W.}},
  \bauthor{\bsnm{Li},~\bfnm{Yingzhen}\binits{Y.}} \AND
  \bauthor{\bsnm{Hern\'{a}ndez-Lobato},~\bfnm{Jos\'{e}~Miguel}\binits{J.~M.}}
(\byear{2019}).
\btitle{Meta-learning For stochastic gradient {MCMC}}.
In \bbooktitle{International Conference on Learning Representations}.
\end{binproceedings}
\endbibitem

\bibitem[\protect\citeauthoryear{Goodfellow, Bengio and
  Courville}{2016}]{goodfellow2016}
\begin{bbook}[author]
\bauthor{\bsnm{Goodfellow},~\bfnm{Ian}\binits{I.}},
  \bauthor{\bsnm{Bengio},~\bfnm{Yoshua}\binits{Y.}} \AND
  \bauthor{\bsnm{Courville},~\bfnm{Aaron}\binits{A.}}
(\byear{2016}).
\btitle{Deep learning}.
\bpublisher{MIT press}.
\end{bbook}
\endbibitem

\bibitem[\protect\citeauthoryear{Graf and Luschgy}{2007}]{graf2007}
\begin{bbook}[author]
\bauthor{\bsnm{Graf},~\bfnm{Siegfried}\binits{S.}} \AND
  \bauthor{\bsnm{Luschgy},~\bfnm{Harald}\binits{H.}}
(\byear{2007}).
\btitle{Foundations of quantization for probability distributions}.
\bpublisher{Springer}.
\end{bbook}
\endbibitem

\bibitem[\protect\citeauthoryear{Gretton et~al.}{2012}]{gretton2012}
\begin{barticle}[author]
\bauthor{\bsnm{Gretton},~\bfnm{Arthur}\binits{A.}},
  \bauthor{\bsnm{Borgwardt},~\bfnm{Karsten~M}\binits{K.~M.}},
  \bauthor{\bsnm{Rasch},~\bfnm{Malte~J}\binits{M.~J.}},
  \bauthor{\bsnm{Sch{\"o}lkopf},~\bfnm{Bernhard}\binits{B.}} \AND
  \bauthor{\bsnm{Smola},~\bfnm{Alexander}\binits{A.}}
(\byear{2012}).
\btitle{A kernel two-sample test}.
\bjournal{Journal of Machine Learning Research}
\bvolume{13}
\bpages{723--773}.
\end{barticle}
\endbibitem

\bibitem[\protect\citeauthoryear{Gu, Ghahramani and Turner}{2015}]{gu2015}
\begin{bincollection}[author]
\bauthor{\bsnm{Gu},~\bfnm{Shixiang~(Shane)}\binits{S.~S.}},
  \bauthor{\bsnm{Ghahramani},~\bfnm{Zoubin}\binits{Z.}} \AND
  \bauthor{\bsnm{Turner},~\bfnm{Richard~E}\binits{R.~E.}}
(\byear{2015}).
\btitle{Neural adaptive sequential {M}onte {C}arlo}.
In \bbooktitle{Advances in Neural Information Processing Systems 28}
\bpages{2629--2637}.
\end{bincollection}
\endbibitem

\bibitem[\protect\citeauthoryear{Hastie, Tibshirani and
  Friedman}{2016}]{hastie2016}
\begin{bbook}[author]
\bauthor{\bsnm{Hastie},~\bfnm{Trevor}\binits{T.}},
  \bauthor{\bsnm{Tibshirani},~\bfnm{Robert}\binits{R.}} \AND
  \bauthor{\bsnm{Friedman},~\bfnm{Jerome}\binits{J.}}
(\byear{2016}).
\btitle{The elements of statistical learning: data mining, inference and
  prediction},
\bedition{2nd} ed.
\bpublisher{Springer}.
\end{bbook}
\endbibitem

\bibitem[\protect\citeauthoryear{Hastings}{1970}]{hastings1970}
\begin{barticle}[author]
\bauthor{\bsnm{Hastings},~\bfnm{W.~K.}\binits{W.~K.}}
(\byear{1970}).
\btitle{Monte {C}arlo sampling methods using {M}arkov chains and their
  applications}.
\bjournal{Biometrika}
\bvolume{57}
\bpages{97--109}.
\end{barticle}
\endbibitem

\bibitem[\protect\citeauthoryear{Hecht-Nielsen}{1990}]{nielsen1990}
\begin{bincollection}[author]
\bauthor{\bsnm{Hecht-Nielsen},~\bfnm{Robert}\binits{R.}}
(\byear{1990}).
\btitle{On the algebraic structure of feedforward network weight spaces}.
In \bbooktitle{Advanced Neural Computers}
\bpages{129--135}.
\end{bincollection}
\endbibitem

\bibitem[\protect\citeauthoryear{Hornik}{1991}]{hornik1991}
\begin{barticle}[author]
\bauthor{\bsnm{Hornik},~\bfnm{Kurt}\binits{K.}}
(\byear{1991}).
\btitle{Approximation capabilities of multilayer feedforward networks}.
\bjournal{Neural Networks}
\bvolume{4}
\bpages{251--257}.
\end{barticle}
\endbibitem

\bibitem[\protect\citeauthoryear{Horst, Hill and Gorman}{2020}]{horst2020}
\begin{bmanual}[author]
\bauthor{\bsnm{Horst},~\bfnm{Allison~Marie}\binits{A.~M.}},
  \bauthor{\bsnm{Hill},~\bfnm{Alison~Presmanes}\binits{A.~P.}} \AND
  \bauthor{\bsnm{Gorman},~\bfnm{Kristen~B}\binits{K.~B.}}
(\byear{2020}).
\btitle{palmerpenguins: {P}almer {A}rchipelago ({A}ntarctica) penguin data}
\bnote{R package version 0.1.0}.
\end{bmanual}
\endbibitem

\bibitem[\protect\citeauthoryear{Howard et~al.}{2017}]{howard2017}
\begin{barticle}[author]
\bauthor{\bsnm{Howard},~\bfnm{Andrew~G}\binits{A.~G.}},
  \bauthor{\bsnm{Zhu},~\bfnm{Menglong}\binits{M.}},
  \bauthor{\bsnm{Chen},~\bfnm{Bo}\binits{B.}},
  \bauthor{\bsnm{Kalenichenko},~\bfnm{Dmitry}\binits{D.}},
  \bauthor{\bsnm{Wang},~\bfnm{Weijun}\binits{W.}},
  \bauthor{\bsnm{Weyand},~\bfnm{Tobias}\binits{T.}},
  \bauthor{\bsnm{Andreetto},~\bfnm{Marco}\binits{M.}} \AND
  \bauthor{\bsnm{Adam},~\bfnm{Hartwig}\binits{H.}}
(\byear{2017}).
\btitle{Mobilenets: efficient convolutional neural networks for mobile vision
  applications}.
\bjournal{arXiv}.
\end{barticle}
\endbibitem

\bibitem[\protect\citeauthoryear{Hu, Zagoruyko and Komodakis}{2019}]{hu2019}
\begin{barticle}[author]
\bauthor{\bsnm{Hu},~\bfnm{Shell~Xu}\binits{S.~X.}},
  \bauthor{\bsnm{Zagoruyko},~\bfnm{Sergey}\binits{S.}} \AND
  \bauthor{\bsnm{Komodakis},~\bfnm{Nikos}\binits{N.}}
(\byear{2019}).
\btitle{Exploring weight symmetry in deep neural networks}.
\bjournal{Computer Vision and Image Understanding}
\bvolume{187}
\bpages{102786}.
\end{barticle}
\endbibitem

\bibitem[\protect\citeauthoryear{Huang et~al.}{2019}]{huang2019}
\begin{binproceedings}[author]
\bauthor{\bsnm{Huang},~\bfnm{Chin-Wei}\binits{C.-W.}},
  \bauthor{\bsnm{Sankaran},~\bfnm{Kris}\binits{K.}},
  \bauthor{\bsnm{Dhekane},~\bfnm{Eeshan}\binits{E.}},
  \bauthor{\bsnm{Lacoste},~\bfnm{Alexandre}\binits{A.}} \AND
  \bauthor{\bsnm{Courville},~\bfnm{Aaron}\binits{A.}}
(\byear{2019}).
\btitle{Hierarchical importance weighted autoencoders}.
In \bbooktitle{Proceedings of the 36th International Conference on Machine
  Learning}
\bvolume{97}
\bpages{2869--2878}.
\end{binproceedings}
\endbibitem

\bibitem[\protect\citeauthoryear{Iandola et~al.}{2016}]{iandola2016}
\begin{barticle}[author]
\bauthor{\bsnm{Iandola},~\bfnm{Forrest~N}\binits{F.~N.}},
  \bauthor{\bsnm{Han},~\bfnm{Song}\binits{S.}},
  \bauthor{\bsnm{Moskewicz},~\bfnm{Matthew~W}\binits{M.~W.}},
  \bauthor{\bsnm{Ashraf},~\bfnm{Khalid}\binits{K.}},
  \bauthor{\bsnm{Dally},~\bfnm{William~J}\binits{W.~J.}} \AND
  \bauthor{\bsnm{Keutzer},~\bfnm{Kurt}\binits{K.}}
(\byear{2016}).
\btitle{Squeeze{N}et: {A}lex{N}et-level accuracy with 50x fewer parameters and
  <0.5MB model size}.
\bjournal{arXiv}.
\end{barticle}
\endbibitem

\bibitem[\protect\citeauthoryear{Izmailov et~al.}{2020}]{izmailov2020}
\begin{binproceedings}[author]
\bauthor{\bsnm{Izmailov},~\bfnm{Pavel}\binits{P.}},
  \bauthor{\bsnm{Maddox},~\bfnm{Wesley~J.}\binits{W.~J.}},
  \bauthor{\bsnm{Kirichenko},~\bfnm{Polina}\binits{P.}},
  \bauthor{\bsnm{Garipov},~\bfnm{Timur}\binits{T.}},
  \bauthor{\bsnm{Vetrov},~\bfnm{Dmitry}\binits{D.}} \AND
  \bauthor{\bsnm{Wilson},~\bfnm{Andrew~Gordon}\binits{A.~G.}}
(\byear{2020}).
\btitle{Subspace inference for {B}ayesian deep learning}.
In \bbooktitle{Proceedings of The 35th Uncertainty in Artificial Intelligence
  Conference}
\bvolume{115}
\bpages{1169--1179}.
\end{binproceedings}
\endbibitem

\bibitem[\protect\citeauthoryear{Jarrett et~al.}{2009}]{jarrett2009}
\begin{binproceedings}[author]
\bauthor{\bsnm{Jarrett},~\bfnm{K.}\binits{K.}},
  \bauthor{\bsnm{Kavukcuoglu},~\bfnm{K.}\binits{K.}},
  \bauthor{\bsnm{Ranzato},~\bfnm{M.}\binits{M.}} \AND
  \bauthor{\bsnm{LeCun},~\bfnm{Y.}\binits{Y.}}
(\byear{2009}).
\btitle{What is the best multi-stage architecture for object recognition?}
In \bbooktitle{IEEE 12th International Conference on Computer Vision}
\bpages{2146-2153}.
\end{binproceedings}
\endbibitem

\bibitem[\protect\citeauthoryear{Jaynes}{1968}]{jaynes1968}
\begin{barticle}[author]
\bauthor{\bsnm{Jaynes},~\bfnm{E.~T.}\binits{E.~T.}}
(\byear{1968}).
\btitle{Prior probabilities}.
\bjournal{IEEE Transactions on Systems Science and Cybernetics}
\bvolume{4}
\bpages{227--241}.
\end{barticle}
\endbibitem

\bibitem[\protect\citeauthoryear{Jeffreys}{1962}]{jeffreys1962}
\begin{bbook}[author]
\bauthor{\bsnm{Jeffreys},~\bfnm{Harold}\binits{H.}}
(\byear{1962}).
\btitle{The theory of probability},
\bedition{3rd} ed.
\bpublisher{OUP Oxford}.
\end{bbook}
\endbibitem

\bibitem[\protect\citeauthoryear{Johndrow, Pillai and
  Smith}{2020}]{johndrow2020}
\begin{barticle}[author]
\bauthor{\bsnm{Johndrow},~\bfnm{James~E.}\binits{J.~E.}},
  \bauthor{\bsnm{Pillai},~\bfnm{Natesh~S.}\binits{N.~S.}} \AND
  \bauthor{\bsnm{Smith},~\bfnm{Aaron}\binits{A.}}
(\byear{2020}).
\btitle{No free lunch for approximate {MCMC}}.
\bjournal{arXiv}.
\end{barticle}
\endbibitem

\bibitem[\protect\citeauthoryear{Kass et~al.}{1998}]{kass1998}
\begin{barticle}[author]
\bauthor{\bsnm{Kass},~\bfnm{Robert~E.}\binits{R.~E.}},
  \bauthor{\bsnm{Carlin},~\bfnm{Bradley~P.}\binits{B.~P.}},
  \bauthor{\bsnm{Gelman},~\bfnm{Andrew}\binits{A.}} \AND
  \bauthor{\bsnm{Neal},~\bfnm{Radford~M.}\binits{R.~M.}}
(\byear{1998}).
\btitle{Markov chain {M}onte {C}arlo in practice: a roundtable discussion}.
\bjournal{The American Statistician}
\bvolume{52}
\bpages{93--100}.
\end{barticle}
\endbibitem

\bibitem[\protect\citeauthoryear{Krizhevsky, Sutskever and
  Hinton}{2012}]{krizhevsky2012}
\begin{bincollection}[author]
\bauthor{\bsnm{Krizhevsky},~\bfnm{Alex}\binits{A.}},
  \bauthor{\bsnm{Sutskever},~\bfnm{Ilya}\binits{I.}} \AND
  \bauthor{\bsnm{Hinton},~\bfnm{Geoffrey~E}\binits{G.~E.}}
(\byear{2012}).
\btitle{Image{N}et classification with deep convolutional neural networks}.
In \bbooktitle{Advances in Neural Information Processing Systems 25}
\bpages{1097--1105}.
\end{bincollection}
\endbibitem

\bibitem[\protect\citeauthoryear{Lee}{2000}]{lee2000}
\begin{barticle}[author]
\bauthor{\bsnm{Lee},~\bfnm{H.~K.~H.}\binits{H.~K.~H.}}
(\byear{2000}).
\btitle{Consistency of posterior distributions for neural networks}.
\bjournal{Neural Networks}
\bvolume{13}
\bpages{629--642}.
\end{barticle}
\endbibitem

\bibitem[\protect\citeauthoryear{Lee}{2003}]{lee2003}
\begin{barticle}[author]
\bauthor{\bsnm{Lee},~\bfnm{Herbert K.~H.}\binits{H.~K.~H.}}
(\byear{2003}).
\btitle{A noninformative prior for neural networks}.
\bjournal{Machine Learning}
\bvolume{50}
\bpages{197--212}.
\end{barticle}
\endbibitem

\bibitem[\protect\citeauthoryear{Lee}{2004}]{lee2004}
\begin{bincollection}[author]
\bauthor{\bsnm{Lee},~\bfnm{Herbert~KH}\binits{H.~K.}}
(\byear{2004}).
\btitle{Priors for neural networks}.
In \bbooktitle{Classification, Clustering, and Data Mining Applications}
(\beditor{\bfnm{David}\binits{D.}~\bsnm{Banks}},
  \beditor{\bfnm{Frederick~R.}\binits{F.~R.}~\bsnm{McMorris}},
  \beditor{\bfnm{Phipps}\binits{P.}~\bsnm{Arabie}} \AND
  \beditor{\bfnm{Wolfgang}\binits{W.}~\bsnm{Gaul}}, eds.)
\bpages{141--150}.
\end{bincollection}
\endbibitem

\bibitem[\protect\citeauthoryear{Lee}{2005}]{lee2005}
\begin{binproceedings}[author]
\bauthor{\bsnm{Lee},~\bfnm{Herbert~KH}\binits{H.~K.}}
(\byear{2005}).
\btitle{Neural networks and default priors}.
In \bbooktitle{Proceedings of the American Statistical Association, Section on
  Bayesian Statistical Science}.
\end{binproceedings}
\endbibitem

\bibitem[\protect\citeauthoryear{Lee}{2007}]{lee2007}
\begin{barticle}[author]
\bauthor{\bsnm{Lee},~\bfnm{Herbert~KH}\binits{H.~K.}}
(\byear{2007}).
\btitle{Default priors for neural network classification}.
\bjournal{Journal of Classification}
\bvolume{24}
\bpages{53--70}.
\end{barticle}
\endbibitem

\bibitem[\protect\citeauthoryear{Lu et~al.}{2017}]{lu2017}
\begin{bincollection}[author]
\bauthor{\bsnm{Lu},~\bfnm{Zhou}\binits{Z.}},
  \bauthor{\bsnm{Pu},~\bfnm{Hongming}\binits{H.}},
  \bauthor{\bsnm{Wang},~\bfnm{Feicheng}\binits{F.}},
  \bauthor{\bsnm{Hu},~\bfnm{Zhiqiang}\binits{Z.}} \AND
  \bauthor{\bsnm{Wang},~\bfnm{Liwei}\binits{L.}}
(\byear{2017}).
\btitle{The expressive power of neural networks: a view from the width}.
In \bbooktitle{Advances in Neural Information Processing Systems 30}
\bpages{6231--6239}.
\end{bincollection}
\endbibitem

\bibitem[\protect\citeauthoryear{Ma, Foti and Fox}{2017}]{ma2017}
\begin{binproceedings}[author]
\bauthor{\bsnm{Ma},~\bfnm{Yi-An}\binits{Y.-A.}},
  \bauthor{\bsnm{Foti},~\bfnm{Nicholas~J.}\binits{N.~J.}} \AND
  \bauthor{\bsnm{Fox},~\bfnm{Emily~B.}\binits{E.~B.}}
(\byear{2017}).
\btitle{Stochastic gradient {MCMC} methods for hidden {M}arkov models}.
In \bbooktitle{Proceedings of the 34th International Conference on Machine
  Learning}
\bvolume{70}
\bpages{2265--2274}.
\end{binproceedings}
\endbibitem

\bibitem[\protect\citeauthoryear{MacKay}{1995}]{mackay1995}
\begin{bincollection}[author]
\bauthor{\bsnm{MacKay},~\bfnm{David~JC}\binits{D.~J.}}
(\byear{1995}).
\btitle{Developments in probabilistic modelling with neural networks—ensemble
  learning}.
In \bbooktitle{Neural Networks: Artificial Intelligence and Industrial
  Applications}
\bpages{191--198}.
\end{bincollection}
\endbibitem

\bibitem[\protect\citeauthoryear{Maddison et~al.}{2015}]{maddison2015}
\begin{binproceedings}[author]
\bauthor{\bsnm{Maddison},~\bfnm{Chris~J}\binits{C.~J.}},
  \bauthor{\bsnm{Huang},~\bfnm{Aja}\binits{A.}},
  \bauthor{\bsnm{Sutskever},~\bfnm{Ilya}\binits{I.}} \AND
  \bauthor{\bsnm{Silver},~\bfnm{David}\binits{D.}}
(\byear{2015}).
\btitle{Move evaluation in {G}o using deep convolutional neural networks}.
In \bbooktitle{International Conference on Learning Representations}.
\end{binproceedings}
\endbibitem

\bibitem[\protect\citeauthoryear{Mandt, Hoffman and Blei}{2017}]{mandt2017}
\begin{barticle}[author]
\bauthor{\bsnm{Mandt},~\bfnm{Stephan}\binits{S.}},
  \bauthor{\bsnm{Hoffman},~\bfnm{Matthew~D.}\binits{M.~D.}} \AND
  \bauthor{\bsnm{Blei},~\bfnm{David~M.}\binits{D.~M.}}
(\byear{2017}).
\btitle{Stochastic gradient descent as approximate {B}ayesian inference}.
\bjournal{Journal of Machine Learning Research}
\bvolume{18}
\bpages{1--35}.
\end{barticle}
\endbibitem

\bibitem[\protect\citeauthoryear{Metropolis et~al.}{1953}]{metropolis1953}
\begin{barticle}[author]
\bauthor{\bsnm{Metropolis},~\bfnm{Nicholas}\binits{N.}},
  \bauthor{\bsnm{Rosenbluth},~\bfnm{Arianna~W}\binits{A.~W.}},
  \bauthor{\bsnm{Rosenbluth},~\bfnm{Marshall~N}\binits{M.~N.}},
  \bauthor{\bsnm{Teller},~\bfnm{Augusta~H}\binits{A.~H.}} \AND
  \bauthor{\bsnm{Teller},~\bfnm{Edward}\binits{E.}}
(\byear{1953}).
\btitle{Equation of state calculations by fast computing machines}.
\bjournal{The journal of Chemical Physics}
\bvolume{21}
\bpages{1087--1092}.
\end{barticle}
\endbibitem

\bibitem[\protect\citeauthoryear{Minsky and Papert}{1988}]{minsky1988}
\begin{bbook}[author]
\bauthor{\bsnm{Minsky},~\bfnm{Marvin~L}\binits{M.~L.}} \AND
  \bauthor{\bsnm{Papert},~\bfnm{Seymour~A}\binits{S.~A.}}
(\byear{1988}).
\btitle{Perceptrons: expanded edition}.
\bpublisher{MIT press}.
\end{bbook}
\endbibitem

\bibitem[\protect\citeauthoryear{Moore}{2016}]{moore2016}
\begin{binproceedings}[author]
\bauthor{\bsnm{Moore},~\bfnm{David~A}\binits{D.~A.}}
(\byear{2016}).
\btitle{Symmetrized variational inference}.
In \bbooktitle{NIPS Workshop on Advances in Approximate {B}ayesian Inference}.
\end{binproceedings}
\endbibitem

\bibitem[\protect\citeauthoryear{Nair and Hinton}{2009}]{nair2009}
\begin{bincollection}[author]
\bauthor{\bsnm{Nair},~\bfnm{Vinod}\binits{V.}} \AND
  \bauthor{\bsnm{Hinton},~\bfnm{Geoffrey~E}\binits{G.~E.}}
(\byear{2009}).
\btitle{3{D} object recognition with deep belief nets}.
In \bbooktitle{Advances in Neural Information Processing Systems 22}
\bpages{1339--1347}.
\end{bincollection}
\endbibitem

\bibitem[\protect\citeauthoryear{Nalisnick}{2018}]{nalisnick2018}
\begin{bphdthesis}[author]
\bauthor{\bsnm{Nalisnick},~\bfnm{Eric~Thomas}\binits{E.~T.}}
(\byear{2018}).
\btitle{On priors for {B}ayesian neural networks},
\btype{PhD thesis},
\bpublisher{UC Irvine}.
\end{bphdthesis}
\endbibitem

\bibitem[\protect\citeauthoryear{Neal}{2011}]{neal2011}
\begin{binbook}[author]
\bauthor{\bsnm{Neal},~\bfnm{R.~M.}\binits{R.~M.}}
(\byear{2011}).
\btitle{{MCMC} Using {H}amiltonian dynamics}
In \bbooktitle{Handbook of Markov Chain Monte Carlo}
\bchapter{5}.
\bpublisher{CRC Press}.
\end{binbook}
\endbibitem

\bibitem[\protect\citeauthoryear{Nemeth and Sherlock}{2018}]{nemeth2018}
\begin{barticle}[author]
\bauthor{\bsnm{Nemeth},~\bfnm{Christopher}\binits{C.}} \AND
  \bauthor{\bsnm{Sherlock},~\bfnm{Chris}\binits{C.}}
(\byear{2018}).
\btitle{Merging {MCMC} subposteriors through {G}aussian-process
  approximations}.
\bjournal{Bayesian Analysis}
\bvolume{13}
\bpages{507--530}.
\end{barticle}
\endbibitem

\bibitem[\protect\citeauthoryear{Nwankpa et~al.}{2018}]{nwankpa2018}
\begin{barticle}[author]
\bauthor{\bsnm{Nwankpa},~\bfnm{Chigozie}\binits{C.}},
  \bauthor{\bsnm{Ijomah},~\bfnm{Winifred}\binits{W.}},
  \bauthor{\bsnm{Gachagan},~\bfnm{Anthony}\binits{A.}} \AND
  \bauthor{\bsnm{Marshall},~\bfnm{Stephen}\binits{S.}}
(\byear{2018}).
\btitle{Activation functions: comparison of trends in practice and research for
  deep learning}.
\bjournal{arXiv}.
\end{barticle}
\endbibitem

\bibitem[\protect\citeauthoryear{Ong, Nott and Smith}{2018}]{ong2018}
\begin{barticle}[author]
\bauthor{\bsnm{Ong},~\bfnm{Victor M.~H.}\binits{V.~M.~H.}},
  \bauthor{\bsnm{Nott},~\bfnm{David~J.}\binits{D.~J.}} \AND
  \bauthor{\bsnm{Smith},~\bfnm{Michael~S.}\binits{M.~S.}}
(\byear{2018}).
\btitle{Gaussian variational approximation with a factor covariance structure}.
\bjournal{Journal of Computational and Graphical Statistics}
\bvolume{27}
\bpages{465--478}.
\end{barticle}
\endbibitem

\bibitem[\protect\citeauthoryear{Pearce et~al.}{2019}]{pearce2019}
\begin{binproceedings}[author]
\bauthor{\bsnm{Pearce},~\bfnm{Tim}\binits{T.}},
  \bauthor{\bsnm{Zaki},~\bfnm{Mohamed}\binits{M.}},
  \bauthor{\bsnm{Brintrup},~\bfnm{Alexandra}\binits{A.}} \AND
  \bauthor{\bsnm{Neely},~\bfnm{Andy}\binits{A.}}
(\byear{2019}).
\btitle{Expressive priors in {B}ayesian neural networks: kernel combinations
  and periodic functions}.
In \bbooktitle{Proceedings of the 35th Conference on Uncertainty in Artificial
  Intelligence}.
\end{binproceedings}
\endbibitem

\bibitem[\protect\citeauthoryear{Polson and Sokolov}{2017}]{polson2017}
\begin{barticle}[author]
\bauthor{\bsnm{Polson},~\bfnm{Nicholas~G.}\binits{N.~G.}} \AND
  \bauthor{\bsnm{Sokolov},~\bfnm{Vadim}\binits{V.}}
(\byear{2017}).
\btitle{Deep learning: a {B}ayesian perspective}.
\bjournal{Bayesian Analysis}
\bvolume{12}
\bpages{1275--1304}.
\end{barticle}
\endbibitem

\bibitem[\protect\citeauthoryear{Pourzanjani, Jiang and
  Petzold}{2017}]{pourzanjani2017}
\begin{binproceedings}[author]
\bauthor{\bsnm{Pourzanjani},~\bfnm{Arya~A}\binits{A.~A.}},
  \bauthor{\bsnm{Jiang},~\bfnm{Richard~M}\binits{R.~M.}} \AND
  \bauthor{\bsnm{Petzold},~\bfnm{Linda~R}\binits{L.~R.}}
(\byear{2017}).
\btitle{Improving the identifiability of neural networks for {B}ayesian
  inference}.
In \bbooktitle{NIPS Workshop on Bayesian Deep Learning}.
\end{binproceedings}
\endbibitem

\bibitem[\protect\citeauthoryear{Quiroz et~al.}{2019}]{quiroz2019}
\begin{barticle}[author]
\bauthor{\bsnm{Quiroz},~\bfnm{Matias}\binits{M.}},
  \bauthor{\bsnm{Kohn},~\bfnm{Robert}\binits{R.}},
  \bauthor{\bsnm{Villani},~\bfnm{Mattias}\binits{M.}} \AND
  \bauthor{\bsnm{Tran},~\bfnm{Minh-Ngoc}\binits{M.-N.}}
(\byear{2019}).
\btitle{Speeding Up {MCMC} by efficient data subsampling}.
\bjournal{Journal of the American Statistical Association}
\bvolume{114}
\bpages{831--843}.
\end{barticle}
\endbibitem

\bibitem[\protect\citeauthoryear{Ranganath, Tran and
  Blei}{2016}]{ranganath2016}
\begin{binproceedings}[author]
\bauthor{\bsnm{Ranganath},~\bfnm{Rajesh}\binits{R.}},
  \bauthor{\bsnm{Tran},~\bfnm{Dustin}\binits{D.}} \AND
  \bauthor{\bsnm{Blei},~\bfnm{David}\binits{D.}}
(\byear{2016}).
\btitle{Hierarchical variational models}.
In \bbooktitle{Proceedings of The 33rd International Conference on Machine
  Learning}
\bvolume{48}
\bpages{324--333}.
\end{binproceedings}
\endbibitem

\bibitem[\protect\citeauthoryear{Robert et~al.}{2018}]{robert2018}
\begin{barticle}[author]
\bauthor{\bsnm{Robert},~\bfnm{Christian~P.}\binits{C.~P.}},
  \bauthor{\bsnm{Elvira},~\bfnm{Víctor}\binits{V.}},
  \bauthor{\bsnm{Tawn},~\bfnm{Nick}\binits{N.}} \AND
  \bauthor{\bsnm{Wu},~\bfnm{Changye}\binits{C.}}
(\byear{2018}).
\btitle{Accelerating {MCMC} algorithms}.
\bjournal{Wiley Interdisciplinary Reviews: Computational Statistics}
\bvolume{10}
\bpages{e1435}.
\end{barticle}
\endbibitem

\bibitem[\protect\citeauthoryear{Rosenblatt}{1958}]{rosenblatt1958}
\begin{barticle}[author]
\bauthor{\bsnm{Rosenblatt},~\bfnm{Frank}\binits{F.}}
(\byear{1958}).
\btitle{The perceptron: a probabilistic model for information storage and
  organization in the brain}.
\bjournal{Psychological review}
\bvolume{65}
\bpages{386}.
\end{barticle}
\endbibitem

\bibitem[\protect\citeauthoryear{Rudolf and Schweizer}{2018}]{rudolf2018}
\begin{barticle}[author]
\bauthor{\bsnm{Rudolf},~\bfnm{Daniel}\binits{D.}} \AND
  \bauthor{\bsnm{Schweizer},~\bfnm{Nikolaus}\binits{N.}}
(\byear{2018}).
\btitle{Perturbation theory for {M}arkov chains via {W}asserstein distance}.
\bjournal{Bernoulli}
\bvolume{24}
\bpages{2610--2639}.
\end{barticle}
\endbibitem

\bibitem[\protect\citeauthoryear{Sargent, Hodges and
  Carlin}{2000}]{sargent2000}
\begin{barticle}[author]
\bauthor{\bsnm{Sargent},~\bfnm{Daniel~J.}\binits{D.~J.}},
  \bauthor{\bsnm{Hodges},~\bfnm{James~S.}\binits{J.~S.}} \AND
  \bauthor{\bsnm{Carlin},~\bfnm{Bradley~P.}\binits{B.~P.}}
(\byear{2000}).
\btitle{Structured {M}arkov chain {M}onte {C}arlo}.
\bjournal{Journal of Computational and Graphical Statistics}
\bvolume{9}
\bpages{217--234}.
\end{barticle}
\endbibitem

\bibitem[\protect\citeauthoryear{Seita et~al.}{2018}]{seita2018}
\begin{binproceedings}[author]
\bauthor{\bsnm{Seita},~\bfnm{Daniel}\binits{D.}},
  \bauthor{\bsnm{Pan},~\bfnm{Xinlei}\binits{X.}},
  \bauthor{\bsnm{Chen},~\bfnm{Haoyu}\binits{H.}} \AND
  \bauthor{\bsnm{Canny},~\bfnm{John}\binits{J.}}
(\byear{2018}).
\btitle{An efficient minibatch acceptance test for {M}etropolis-{H}astings}.
In \bbooktitle{Proceedings of the Twenty-Seventh International Joint Conference
  on Artificial Intelligence}
\bpages{5359--5363}.
\end{binproceedings}
\endbibitem

\bibitem[\protect\citeauthoryear{Sen, Papamarkou and Dunson}{2020}]{sen2020}
\begin{barticle}[author]
\bauthor{\bsnm{Sen},~\bfnm{Deborshee}\binits{D.}},
  \bauthor{\bsnm{Papamarkou},~\bfnm{Theodore}\binits{T.}} \AND
  \bauthor{\bsnm{Dunson},~\bfnm{David}\binits{D.}}
(\byear{2020}).
\btitle{Bayesian neural networks and dimensionality reduction}.
\bjournal{arXiv}.
\end{barticle}
\endbibitem

\bibitem[\protect\citeauthoryear{Simpson et~al.}{2017}]{simpson2017}
\begin{barticle}[author]
\bauthor{\bsnm{Simpson},~\bfnm{Daniel}\binits{D.}},
  \bauthor{\bsnm{Rue},~\bfnm{H\r{a}vard}\binits{H.}},
  \bauthor{\bsnm{Riebler},~\bfnm{Andrea}\binits{A.}},
  \bauthor{\bsnm{Martins},~\bfnm{Thiago~G.}\binits{T.~G.}} \AND
  \bauthor{\bsnm{S{\o}rbye},~\bfnm{Sigrunn~H.}\binits{S.~H.}}
(\byear{2017}).
\btitle{Penalising model component complexity: a principled, practical approach
  to constructing priors}.
\bjournal{Statistical Science}
\bvolume{32}
\bpages{1--28}.
\end{barticle}
\endbibitem

\bibitem[\protect\citeauthoryear{Smith et~al.}{1988}]{smith1988}
\begin{binproceedings}[author]
\bauthor{\bsnm{Smith},~\bfnm{Jack~W}\binits{J.~W.}},
  \bauthor{\bsnm{Everhart},~\bfnm{JE}\binits{J.}},
  \bauthor{\bsnm{Dickson},~\bfnm{WC}\binits{W.}},
  \bauthor{\bsnm{Knowler},~\bfnm{WC}\binits{W.}} \AND
  \bauthor{\bsnm{Johannes},~\bfnm{RS}\binits{R.}}
(\byear{1988}).
\btitle{Using the {ADAP} learning algorithm to forecast the onset of diabetes
  mellitus}.
In \bbooktitle{Proceedings of the Annual Symposium on Computer Application in
  Medical Care}
\bpages{261}.
\end{binproceedings}
\endbibitem

\bibitem[\protect\citeauthoryear{Stephens}{2000}]{stephens2000}
\begin{barticle}[author]
\bauthor{\bsnm{Stephens},~\bfnm{M.}\binits{M.}}
(\byear{2000}).
\btitle{Dealing with label switching in mixture models}.
\bjournal{Journal of the Royal Statistical Society: Series B (Statistical
  Methodology)}
\bvolume{62}
\bpages{795--809}.
\end{barticle}
\endbibitem

\bibitem[\protect\citeauthoryear{Titsias and Ruiz}{2019}]{titsias2019}
\begin{binproceedings}[author]
\bauthor{\bsnm{Titsias},~\bfnm{Michalis~K.}\binits{M.~K.}} \AND
  \bauthor{\bsnm{Ruiz},~\bfnm{Francisco}\binits{F.}}
(\byear{2019}).
\btitle{Unbiased implicit variational inference}.
In \bbooktitle{Proceedings of Machine Learning Research}
\bvolume{89}
\bpages{167--176}.
\end{binproceedings}
\endbibitem

\bibitem[\protect\citeauthoryear{Titterington}{2004}]{titterington2004}
\begin{barticle}[author]
\bauthor{\bsnm{Titterington},~\bfnm{D.~M.}\binits{D.~M.}}
(\byear{2004}).
\btitle{Bayesian methods for neural networks and related models}.
\bjournal{Statistical Science}
\bvolume{19}
\bpages{128--139}.
\end{barticle}
\endbibitem

\bibitem[\protect\citeauthoryear{Truong, Nguyen and Tran}{2018}]{truong2018}
\begin{binproceedings}[author]
\bauthor{\bsnm{Truong},~\bfnm{Thanh-Dat}\binits{T.-D.}},
  \bauthor{\bsnm{Nguyen},~\bfnm{Vinh-Tiep}\binits{V.-T.}} \AND
  \bauthor{\bsnm{Tran},~\bfnm{Minh-Triet}\binits{M.-T.}}
(\byear{2018}).
\btitle{Lightweight Deep Convolutional Network for Tiny Object Recognition}.
In \bbooktitle{Proceedings of the 7th International Conference on Pattern
  Recognition Applications and Methods}
\bpages{675--682}.
\end{binproceedings}
\endbibitem

\bibitem[\protect\citeauthoryear{Vats and Flegal}{2018}]{vats2018b}
\begin{barticle}[author]
\bauthor{\bsnm{Vats},~\bfnm{Dootika}\binits{D.}} \AND
  \bauthor{\bsnm{Flegal},~\bfnm{James~M}\binits{J.~M.}}
(\byear{2018}).
\btitle{Lugsail lag windows and their application to {MCMC}}.
\bjournal{arXiv}.
\end{barticle}
\endbibitem

\bibitem[\protect\citeauthoryear{Vats, Flegal and Jones}{2019}]{vats2019}
\begin{barticle}[author]
\bauthor{\bsnm{Vats},~\bfnm{Dootika}\binits{D.}},
  \bauthor{\bsnm{Flegal},~\bfnm{James~M}\binits{J.~M.}} \AND
  \bauthor{\bsnm{Jones},~\bfnm{Galin~L}\binits{G.~L.}}
(\byear{2019}).
\btitle{Multivariate output analysis for {M}arkov chain {M}onte {C}arlo}.
\bjournal{Biometrika}
\bvolume{106}
\bpages{321--337}.
\end{barticle}
\endbibitem

\bibitem[\protect\citeauthoryear{Vats and Knudson}{2018}]{vats2018a}
\begin{barticle}[author]
\bauthor{\bsnm{Vats},~\bfnm{Dootika}\binits{D.}} \AND
  \bauthor{\bsnm{Knudson},~\bfnm{Christina}\binits{C.}}
(\byear{2018}).
\btitle{Revisiting the {G}elman-{R}ubin diagnostic}.
\bjournal{arXiv}.
\end{barticle}
\endbibitem

\bibitem[\protect\citeauthoryear{Vehtari et~al.}{2019}]{vehtari2019}
\begin{barticle}[author]
\bauthor{\bsnm{Vehtari},~\bfnm{Aki}\binits{A.}},
  \bauthor{\bsnm{Gelman},~\bfnm{Andrew}\binits{A.}},
  \bauthor{\bsnm{Simpson},~\bfnm{Daniel}\binits{D.}},
  \bauthor{\bsnm{Carpenter},~\bfnm{Bob}\binits{B.}} \AND
  \bauthor{\bsnm{Burkner},~\bfnm{Paul-Christian}\binits{P.-C.}}
(\byear{2019}).
\btitle{Rank-normalization, folding, and localization: an improved {R} for
  assessing convergence of {MCMC}}.
\bjournal{arXiv}.
\end{barticle}
\endbibitem

\bibitem[\protect\citeauthoryear{Vladimirova et~al.}{2019}]{vladimirova2019}
\begin{binproceedings}[author]
\bauthor{\bsnm{Vladimirova},~\bfnm{Mariia}\binits{M.}},
  \bauthor{\bsnm{Verbeek},~\bfnm{Jakob}\binits{J.}},
  \bauthor{\bsnm{Mesejo},~\bfnm{Pablo}\binits{P.}} \AND
  \bauthor{\bsnm{Arbel},~\bfnm{Julyan}\binits{J.}}
(\byear{2019}).
\btitle{Understanding priors in {B}ayesian neural networks at the unit level}.
In \bbooktitle{Proceedings of the 36th International Conference on Machine
  Learning}
\bvolume{97}
\bpages{6458--6467}.
\end{binproceedings}
\endbibitem

\bibitem[\protect\citeauthoryear{Welling and Teh}{2011}]{welling2011}
\begin{binproceedings}[author]
\bauthor{\bsnm{Welling},~\bfnm{Max}\binits{M.}} \AND
  \bauthor{\bsnm{Teh},~\bfnm{Yee~Whye}\binits{Y.~W.}}
(\byear{2011}).
\btitle{Bayesian learning via stochastic gradient {L}angevin dynamics}.
In \bbooktitle{Proceedings of the 28th International Conference on
  International Conference on Machine Learning}
\bpages{681--688}.
\end{binproceedings}
\endbibitem

\bibitem[\protect\citeauthoryear{Williams}{1995}]{williams1995}
\begin{barticle}[author]
\bauthor{\bsnm{Williams},~\bfnm{Peter~M.}\binits{P.~M.}}
(\byear{1995}).
\btitle{Bayesian regularization and pruning using a {L}aplace prior}.
\bjournal{Neural Computation}
\bvolume{7}
\bpages{117--143}.
\end{barticle}
\endbibitem

\bibitem[\protect\citeauthoryear{Williams}{2000}]{williams2000}
\begin{bincollection}[author]
\bauthor{\bsnm{Williams},~\bfnm{Christopher K.~I.}\binits{C.~K.~I.}}
(\byear{2000}).
\btitle{An {MCMC} approach to hierarchical mixture modelling}.
In \bbooktitle{Advances in Neural Information Processing Systems 12}
\bpages{680--686}.
\end{bincollection}
\endbibitem

\bibitem[\protect\citeauthoryear{Wilson and Izmailov}{2020}]{wilson2020}
\begin{barticle}[author]
\bauthor{\bsnm{Wilson},~\bfnm{Andrew~Gordon}\binits{A.~G.}} \AND
  \bauthor{\bsnm{Izmailov},~\bfnm{Pavel}\binits{P.}}
(\byear{2020}).
\btitle{Bayesian deep learning and a probabilistic perspective of
  generalization}.
\bjournal{ar{X}iv}.
\end{barticle}
\endbibitem

\bibitem[\protect\citeauthoryear{Zhang et~al.}{2018a}]{zhang2018a}
\begin{binproceedings}[author]
\bauthor{\bsnm{Zhang},~\bfnm{Guodong}\binits{G.}},
  \bauthor{\bsnm{Sun},~\bfnm{Shengyang}\binits{S.}},
  \bauthor{\bsnm{Duvenaud},~\bfnm{David}\binits{D.}} \AND
  \bauthor{\bsnm{Grosse},~\bfnm{Roger}\binits{R.}}
(\byear{2018}a).
\btitle{Noisy natural gradient as variational inference}.
In \bbooktitle{Proceedings of the 35th International Conference on Machine
  Learning}
\bvolume{80}
\bpages{5852--5861}.
\end{binproceedings}
\endbibitem

\bibitem[\protect\citeauthoryear{Zhang et~al.}{2018b}]{zhang2018b}
\begin{binproceedings}[author]
\bauthor{\bsnm{Zhang},~\bfnm{Xiangyu}\binits{X.}},
  \bauthor{\bsnm{Zhou},~\bfnm{Xinyu}\binits{X.}},
  \bauthor{\bsnm{Lin},~\bfnm{Mengxiao}\binits{M.}} \AND
  \bauthor{\bsnm{Sun},~\bfnm{Jian}\binits{J.}}
(\byear{2018}b).
\btitle{Shufflenet: An extremely efficient convolutional neural network for
  mobile devices}.
In \bbooktitle{Proceedings of the IEEE Conference on Computer Vision and
  Pattern Recognition}
\bpages{6848--6856}.
\end{binproceedings}
\endbibitem

\end{thebibliography}

\end{document}